\documentclass[preprint,12pt,authoryear]{elsarticle}

\usepackage{amssymb}

\usepackage[top = 20mm, bottom = 20mm, left = 25mm, right = 25mm]{geometry}
\usepackage{xcolor}
\usepackage{amsmath}
\usepackage{graphicx}
\usepackage{enumerate}
\usepackage{booktabs}
\usepackage{multirow}
\usepackage{verbatim}
\usepackage{natbib}
\usepackage[colorlinks,
linkcolor=red,
anchorcolor=blue,
citecolor=green
]{hyperref}

\usepackage[noend]{algpseudocode}
\usepackage{algorithmicx,algorithm} 
\usepackage{ntheorem}      

\newtheorem{thm}{\bf Theorem}[section]
\newtheorem{assumption}{Assumption}
\newtheorem{lemma}{Lemma}

\newtheorem{definition}{Definition}
\newtheorem*{Proof}{Proof}

\makeatletter
\newenvironment{breakablealgorithm}
{
	\begin{center}
		\refstepcounter{algorithm}
		\hrule height.8pt depth0pt \kern2pt
		\renewcommand{\caption}[2][\relax]{
			{\raggedright\textbf{Algorithm~\thealgorithm} ##2\par}
			\ifx\relax##1\relax 
			\addcontentsline{loa}{algorithm}{\protect\numberline{\thealgorithm}##2}%
			\else 
			\addcontentsline{loa}{algorithm}{\protect\numberline{\thealgorithm}##1}%
			\fi
			\kern2pt\hrule\kern2pt
		}
	}{
		\kern2pt\hrule\relax
	\end{center}
}
\makeatother

\makeatletter
\def\@author#1{\g@addto@macro\elsauthors{\normalsize%
		\def\baselinestretch{1}%
		\upshape\authorsep#1\unskip\textsuperscript{%
			\ifx\@fnmark\@empty\else\unskip\sep\@fnmark\let\sep=,\fi
			\ifx\@corref\@empty\else\unskip\sep\@corref\let\sep=,\fi
		}%
		\def\authorsep{\unskip,\space}%
		\global\let\@fnmark\@empty
		\global\let\@corref\@empty  
		\global\let\sep\@empty}%
	\@eadauthor={#1}
}
\makeatother

\journal{Journal of Machine Learning Research}

\begin{document}

\begin{frontmatter}



\title{Stability and $L_2$-penalty in Model Averaging}


\author[label1]{Hengkun Zhu}
\ead{hengkunzhu@163.com}
\author[label1]{Guohua Zou }
\ead{ghzou@amss.ac.cn.}

\affiliation[label1]{organization={School of Mathematical Sciences, Capital Normal University},
	city={Beijing},
	postcode={100048},
	country={China} 
}

\begin{abstract}
	Model averaging has received much attention in the past two decades, which 
	integrates available information by averaging over potential models. 
	Although various model averaging methods have been developed, there are few 
	literatures on the theoretical properties of model averaging from the 
	perspective of stability, and the majority of these methods constrain model weights to a simplex. The aim of this paper is to introduce stability from statistical learning theory into model averaging. Thus, we define the 
	stability, asymptotic empirical risk minimizer, generalization, and 
	consistency of model averaging and study the relationship among them. Our 
	results indicate that stability can ensure that model averaging has good 
	generalization performance and consistency under reasonable conditions, 
	where consistency means model averaging estimator can asymptotically minimize the mean squared	 
	prediction error. We also propose a $L_2$-penalty model 
	averaging method without limiting model weights and prove that it has 
	stability and consistency. In order to reduce the impact of tuning 
	parameter selection, we use 10-fold cross-validation to select 
	a candidate set of tuning parameters  and perform a weighted average of 
	the estimators of model weights based on estimation errors. The Monte 
	Carlo simulation and an illustrative application demonstrate the usefulness 
	of the proposed method. 
\end{abstract}


		


\end{frontmatter}


\section{Introduction}
\label{sec1}

	In practical applications, data analysts usually determine a series of models based on exploratory analysis for data and empirical knowledge to describe the relationship between variables of interest and related variables, but how to use these models to produce good results is a more important problem. It is very common to select one model using some data driven criteria, such as AIC\citep{AIC}, BIC\citep{BIC}, $C_p$\citep{Mallows} and FIC\citep{FIC}. However, the properties of the corresponding results depend not only on the selected model, but also on the randomness of the selected model. An alternative to model selection is to make the resultant estimator compromises across a set of competing models. Further, statisticians find that they can usually obtain better and more stable results by combining information from different models. This process of combining multiple models into estimation is known as model averaging. Up to now, there have been a lot of literatures on Bayesian model averaging (BMA) and frequentist model averaging (FMA). \cite{RBMA} reviewed the relevant  literature on BMA. In this paper, we focus on FMA. In the past decades, the model averaging method has been applied in various fields. \cite{App-MA-2009} examined applications of model averaging in tourism research. \cite{App-MA-2011} applied the model averaging method to grain production forecasting of China. \cite{App-MA-2015} reviewed the literature on model averaging with special emphasis on its applications to economics. The key to FMA lies in how to select model weights. The common weight selection methods include: 1) methods based on information criteria, such as smoothed AIC and smoothed BIC in \cite{SAIC}; 2) Mallows model averaging(MMA), proposed by \cite {HMMA}(see also \citealp{WMMA}), modified by \cite{CMMA1} to make it applicable to heteroscedasticity, and improved by \cite{CMMA2} in small sample sizes; 3) FIC criterion, proposed by \cite {FIC}, which is extended to generalized additive partial linear model\citep{MFIC-2011}, the Tobit model with non-zero threshold\citep{MFIC-2012}, and the weighted composite quantile regression\citep{MFIC-2014}; 4) adaptive methods, such as \cite{AM2001} and \cite{AM2005}; 5) OPT method\citep{OPT2011}; 6) cross validation methods, such as jackknife model averaging(JMA)\citep{JMA2012}, \cite{JMA2013}, \cite{CV2016}, \cite{CV2020} and \cite{GMMA}, among others. 
	
	In the learning theory, stability is used to measure algorithm's sensitivity to perturbations in the training set, and is an important tool for analyzing generalization and learnability of algorithms. \cite{SG2002} introduced four kinds of stabilities (hypothesis stability, pointwise hypothesis stability, error stability, and uniform stability), and showed that stability is a sufficient condition for learnability. \cite{WSG2002} introduced several weaker variants of stability, and showed how they are sufficient to obtain generalization bounds for algorithms stable in their sense. \cite{SG2005} and \cite{SG2006} discussed the necessity of stability for learnability under the assumption that uniform convergence is equivalent to learnability. Further, \cite{SLT0} showed that uniform convergence is in fact not necessary for learning in the general learning setting, and stability plays there a key role which has nothing to do with uniform convergence.  In the general learning setting with differential privacy constraint, \cite{SG2016} studied
	some intricate relationships between privacy, stability and learnability.
	
	Although various model averaging methods have been proposed, there are few literatures on their theoretical properties from the perspective of stability, and the majority of these methods are concerned only with whether the resultant estimator leads to a good approximation for the minimum of a given target when the model weights are constrained to a simplex. Considering that there is still a lack of literature on whether stability can ensure good theoretical properties of the model average estimator, and it may be better to approximate the global minimum than the local minimum in some cases, our attempts in this paper are to study stability in model averaging and to answer whether the resultant estimator can lead to a good approximation for the minimum of a given target when the model weights are unrestricted.
	
	For the first attempt, we introduce the concept of stability from 
	statistical learning theory into model averaging. Stability discusses how 
	much the 
	algorithm's output  varies when the sample set changes a little. \cite{SLT0} 
	discussed the relationship among asymptotic empirical risk minimizer 
	(AERM), stability, generalization and consistency, but relevant conclusions 
	cannot be directly applied to model averaging. Therefore, we explore the 
	relevant definitions and conclusions of \cite{SLT0} in model averaging. The 
	results indicate that stability can ensure that model averaging has good 
	generalization performance and consistency under reasonable conditions, 
	where consistency means model averaging estimator can asymptotically minimize the mean squared 
	prediction error (MSPE). For the second one, we find, for MMA and JMA, 
	extreme weights 
	tend to appear with the influence of correlation among models when the 
	model weights are unrestricted, resulting in poor performance of model 
	averaging estimator. Therefore, we should not simply remove the weight 
	constraint and directly use the existing model averaging method. Thus, 
	similar to ridge regression in \cite{RLSEl1970}, we introduce $L_2$-penalty 
	for the weight vector in MMA and JMA and call them Ridge-Mallows model 
	averaging (RMMA) and Ridge-jackknife model averaging (RJMA), respectively. 
	Like Theorem 4.3 in \cite{RLSEl1970}, we discuss the reasonability of 
	introducing $L_2$-penalty. We also prove the stability and consistency of 
	the proposed method, where consistency means model averaging estimator can asymptotically minimize MSPE when 
	the model weights are unrestricted. In the context of shrinkage estimation, 
	\cite{SMA} discussed the impact of tuning parameter selection, 
	and pointed out that the weighted average of shrinkage estimators with 
	different tuning parameters can improve overall stability, predictive 
	performance and standard errors of shrinkage estimators. Hence, like \cite{SMA}, we use 
	10-fold cross-validation to select a candidate set of tuning parameters  and 
	perform a weighted average of the estimators of model weights based on 
	estimation errors.
	
	The remainder of this paper is organised as follows. In Section 
	$\ref{sec2}$, we explain the relevant notations, give the definitions of 
	consistency and stability, and discuss their relationship. In Section $\ref{sec3}$, we 
	propose RMMA and RJMA methods, and prove that they are stable and consistent. 
	Section $\ref{sec4}$ conducts the Monte Carlo simulation experiment. Section 
	$\ref{sec5}$ applies the proposed method to a real data set. Section $\ref{sec6}$ concludes. The proofs of lemmas and theorems 
	are provided in the Appendix.

\section{ Consistency and Stability for Model Averaging}
\label{sec2}

\subsection{ Model Averaging }
\label{sec2-1}

	We assume that $S=\{ z_i=(y_i,x_i^{'})^{'} \in \mathcal{Z},i=1,...,n \}$ is a simple random sample from distribution $\mathcal{D}$, where $y_i$ is the $i$-th observation of response variable, and $x_i$ is the $i$-th observation of covariates. Let $z^*=(y^*,x^{*'})^{'}$ be an observation from distribution $\mathcal{D}$ and independent of $S$.
	
	In model averaging, $M$ approximating models are selected first in order to describe the relationship between response variable and covariates. We assume that the hypothesis spaces of $M$ approximating models are 
	$$\mathcal{H}_m = \big\{ h_m(x^*_m),h_m \in \mathcal{F}_m  \big\},m=1,...,M,$$
	where $x^{*}_m $ consists of some elements of $x^{*}$, and $\mathcal{F}_m$ is a given function set. For example, in MMA, to estimate $E(y^*|x^*)$, we take
	$$ \mathcal{H}_m = \big\{ x^{*'}_m \theta_m ,\theta_m \in R^{ dim(x^{*}_m) } \big\},m=1,...,M, $$
	where $dim( \cdot )$ represents the dimension of the vector. For the $m$-th approximating model, a proper estimation method $A_m$ is selected, and $\hat{h}_m$, the estimator of $h_m$, is obtained based on $S$ and $A_m$. Then, the hypothesis space of model averaging is defined as follows:
	$$ \mathcal{H} = \big\{\hat{h}(x^*,w)= H[w,\hat{h}_1(x^*_1),...,\hat{h}_M(x^*_M)],w \in W  \big\}, $$
	where $W$ is a given weight space, and $H(\cdot)$ is a given function of weight vector and estimators of $M$ approximating models. In MMA, we take
	$$ H[w,\hat{h}_1(x^*_1),...,\hat{h}_M(x^*_M)] = \sum_{m=1}^m w_m  \hat{h}_m(x^*_m) $$
	as the model average estimator of $E(y^*|x^*)$. An important problem with model averaging is the choice of model weights. Here, the estimator $\hat{w}$ of weight vector is obtained based on $S$ and a proper weight selection criterion $A(w)$ that makes $\hat{w}$ be optimal in a certain sense.
	
	The selection of $A_m,m=1,...,M$ and $A(w)$ is closely related to the definition of the loss function. Let $L[\hat{h}(x^*,w),y^*]$ be a real value loss function which is defined in $\mathcal{H} \times \mathcal{Y}$, where $\mathcal{Y}$ is the value space of $y^*$. Then, the risk function is defined as follows:
	$$ F(w,S) = E_{ z^* } \big\{L[\hat{h}(x^*,w),y^*]  \big\}  ,$$
	which is an MSPE under the sample set S and weight vector $w$. 	

\subsection{Related Concepts}
\label{sec2-2}

	In this paper, we mainly discuss whether 
	$F(\hat{w},S)$ can approximate smallest possible risk $inf_{w \in W} 
	F(w,S)$. If $A(w)$ has such a property, we say that $A(w)$ is consistent. For fixed $m$, \cite{SLT0} defined the stability and consistency of $A_m$, and discussed their relationship. Obviously, for model averaging, we need to pay more attention to the stability and consistency of weight selection. We note that relevant conclusions of \cite{SLT0} cannot be directly applied to model averaging because $\mathcal{H}$ depends on $S$. Therefore, we extend the relevant definitions and conclusions to model averaging. The following is the definition of consistency:
	\begin{definition}[Consistency]
		If  there is a sequence of constants $\{\epsilon_{n},n \in N_+ \}$ such that $\epsilon_{n}=o(1)$ and $\hat{w}$ satisfies
		$$ E_{S} \big\{ F(\hat{w},S) - inf_{w \in W} F(w,S)  \big\} = O( \epsilon_{n} )  ,$$
		then $A(w)$ is said to be consistent with rate $\epsilon_{n}$.
		\label{def1}
	\end{definition}
	
	In statistical learning theory, stability concerns how much the algorithm's output varies when $S$ changes a little.``Leave-one-out(Loo)" and ``Replace-one(Ro)" are two common tools used to evaluate stability. Loo considers the change in the algorithm's output after removing an observation from $S$, and Ro considers such a change after replacing an observation in $S$ with an observation that is independent of S. Accordingly, the stability is called Loo stability and Ro stability respectively. Here we will give the  formal definitions of Loo stability and Ro stability. To this end, we first give the definition of algorithm symmetry:
	\begin{definition}[Symmetry]
		If the algorithm's output is not affected by the order of the observations in $S$, the algorithm is symmetric. 
		\label{def2}
	\end{definition}
	
	Now let $S^{ -i }$ be the sample set after removing the $i$-th observation from $S$, $\hat{h}^{-i}_m $ be the estimator of $h_m$ based on $S^{ -i }$ and $A_m$, $\hat{w}^{-i}$ be the estimator  of weight vector based on $S^{ -i }$ and $A(w)$, and $F(w,S^{-i}) = E_{ z^* } \big\{ L[\hat{h}^{-i}(x^*,w),y^*] \big\}  $, where $\hat{h}^{-i}(x^*,w)= H[w,\hat{h}^{-i}_1(x^*_1),...,\hat{h}^{-i}_M(x^*_M)]$. We define Loo stability as follows:
	\begin{definition}[PLoo Stability]
		If  there is a sequence of constants $\{\epsilon_{n},n \in N_+ \}$ such that $\epsilon_{n}=o(1)$ and $A(w)$ satisfies
		$$ \frac{1}{n}\sum_{i=1}^n E_{S} \big[ F(\hat{w},S) - F(\hat{w}^{-i},S^{-i}) \big] =O( \epsilon_{n} )  , $$
		then $A(w)$ is Predicted-Loo(PLoo) stable with rate $\epsilon_{n}$; If $A_m,m=1,...,M$ and $A(w)$ are symmetric, it only has to satisfy 
		$$ E_{S} \big[ F(\hat{w},S) - F(\hat{w}^{-n},S^{-n}) \big] =O( \epsilon_{n} ) .$$
		\label{def3-1}
	\end{definition}
	
	\begin{definition}[FLoo Stability]
		If  there is a sequence of constants $\{\epsilon_{n},n \in N_+ \}$ such that $\epsilon_{n}=o(1)$ and $A(w)$ satisfies
		$$ \frac{1}{n}\sum_{i=1}^n E_{S  } \big\{ L[\hat{h}(x_i,\hat{w}),y_i] - L[\hat{h}^{-i}(x_i,\hat{w}^{-i}),y_i] \big\} =O( \epsilon_{n} )  , $$
		then $A(w)$ is Fitted-Loo(FLoo) stable with rate $\epsilon_{n}$; If $A_m,m=1,...,M$ and $A(w)$ are symmetric, it only has to satisfy 
		$$E_{S  } \big\{ L[\hat{h}(x_n,\hat{w}),y_n] - L[\hat{h}^{-n}(x_n,\hat{w}^{-n}),y_n ]\big\} =O( \epsilon_{n} ).$$
		\label{def3-2}
	\end{definition}
	Let $S^{ i }$ be the sample set S with the $i$-th observation replaced by $z_i^*= (y_i^*,x_i^{*'})^{'}$, $\hat{h}^{i}_m $ be the estimator of $h_m$ based on $S^{ i }$ and $A_m$, and $\hat{w}^{i}$ be the estimator of weight vector based on $S^{ i }$ and $A(w)$, where $z_i^*$ is from distribution $\mathcal{D}$ and independent of $S$. Let $F(w,S^{i}) = E_{ z^*} \big\{ L[\hat{h}^{i}(x^*,w),y^*] \big\} $, then 
	$$ \frac{1}{n}\sum_{i=1}^n E_{S ,z_i^* } \big[ F(\hat{w},S) - 
	F(\hat{w}^{i},S^{i}) \big]=0, $$
	where $\hat{h}^{i}(x^*,w)= H[w,\hat{h}^{i}_1(x^*_1),...,\hat{h}^{i}_M(x^*_M)]$. Therefore, we define Ro stability as follows:
	\begin{definition}[Ro Stability]
		If  there is a sequence of constants $\{\epsilon_{n},n \in N_+ \}$ such that $\epsilon_{n}=o(1)$ and $A(w)$ satisfies
		$$ \frac{1}{n}\sum_{i=1}^n E_{S ,z_i^* } \big\{ L[\hat{h}(x_i,\hat{w}),y_i] - 
		L[\hat{h}^{i}(x_i,\hat{w}^{i}),y_i] \big\}=O(\epsilon_{n})   , $$
		then $A(w)$ is Ro stable with rate $\epsilon_{n}$; If $A_m,m=1,...,M$ and $A(w)$ are symmetric, it only has to satisfy
		$$E_{S ,z_n^*} \big\{ L[\hat{h}(x_n,\hat{w}),y_n] - 
		L[\hat{h}^{n}(x_n,\hat{w}^{n}),y_n]\big\} =O(\epsilon_{n})   .$$
		\label{def4}
	\end{definition}
	
	Before discussing the relationship between stability and consistency, we give the definitions of AERM and generalization. The empirical risk function is defined as follows:
	$$ \hat{F}(w,S) = \frac{1}{n} \sum_{n=1}^n L[\hat{h}(x_i,w),y_i ].$$ 
	\begin{definition}[AERM]
		If  there is a sequence of constants $\{\epsilon_{n},n \in N_+ \}$ such that $\epsilon_{n}=o(1)$ and $A(w)$ satisfies 
		$$  E_{S  } \big[ \hat{F}(\hat{w},S) - inf_{w \in W} \hat{F}(w,S)  \big] =O(\epsilon_{n}) , $$
		then $A(w)$ is an AERM with rate $\epsilon_{n}$.
		\label{def5}
	\end{definition}	
	\cite{SLT1998} proved some theoretical properties of the empirical risk minimization principle. However, when the sample size is small, the empirical risk minimizer tends to produce over-fitting phenomenon. Therefore, the structural risk minimization principle is proposed in \cite{SLT1998}, and the method to satisfy this principle is usually an AERM. \cite{SLT0} also discussed the deficiency of the empirical risk minimization principle and the importance of AERM.
	\begin{definition}[Generalization]
		If  there is a sequence of constants $\{\epsilon_{n},n \in N_+ \}$ such that $\epsilon_{n}=o(1)$ and $A(w)$ satisfies 
		$$  E_{S  } \big[ \hat{F}(\hat{w},S) - F(\hat{w},S)  \big] = O(\epsilon_{n})  , $$
		then $A(w)$  generalizes with rate $\epsilon_{n}$.
		\label{def6}
	\end{definition}
	In statistical learning theory, generalization refers to the performance of the concept learned by models on unknown samples. It can be seen from Definition \ref{def6} that the generalization of $A(w)$ describes the difference between using $\hat{w}$ to fit the training set $S$ and predict the unknown sample.

\subsection{Relationship between Different Concepts}
\label{sec2-3}

	Note that for any $i \in \{1,...,n\}$,
	\begin{align*}
		&  E_{S  ,z_i^*  } \Big\{ L[\hat{h}(x_i,\hat{w}),y_i] - 
		L[\hat{h}^{i}(x_i,\hat{w}^{i}),y_i] \Big\}   \\
		&= E_{S  ,z_i^*  } \Big\{ L[\hat{h}(x_i,\hat{w}),y_i] 
		- L[\hat{h}^{-i}(x_i,\hat{w}^{-i}),y_i ]
		+L[\hat{h}^{-i}(x_i,\hat{w}^{-i}),y_i] -L[\hat{h}^{i}(x_i,\hat{w}^{i}),y_i] \Big\} \\
		&=E_{S  } \Big\{ L[\hat{h}(x_i,\hat{w}),y_i ]
		- L[\hat{h}^{-i}(x_i,\hat{w}^{-i}),y_i]\Big\} 
		+E_{S  } [F(\hat{w}^{-i},S^{-i}) - F(\hat{w},S)],
	\end{align*}
	so we give the following theorem to illustrate the relationship between Loo stability and Ro stability:
	\begin{thm}
		If $A(w)$ has two of FLoo stability, PLoo stability and Ro stability with rate $ \epsilon_{n} $, then it has all three stabilities with rate $ \epsilon_{n} $. 
		\label{thm:3-1}
	\end{thm}
	\cite{SLT0} emphasized that Ro stability and Loo stability are in general incomparable notions, but Theorem \ref{thm:3-1} shows that they are closely related. 
	
	By definitions of generalization and Ro stability, we have
	\begin{align*}
		&E_{S  } [ \hat{F}(\hat{w},S) - F(\hat{w},S)  ]  \\
		&=E_{S  ,z_1^*,...,z_n^*   } \Big\{ \frac{1}{n} \sum_{n=1}^n L[ 	\hat{h}(x_i,\hat{w}),y_i ] - \frac{1}{n}  \sum_{n=1}^n  L [\hat{h}(x_i^*,\hat{w}),y_i^* ] \Big \} \\
		& =\frac{1}{n} \sum_{n=1}^n E_{S  ,z_i^*  } \Big\{ L[\hat{h}(x_i,\hat{w}),y_i] - 
		L[\hat{h}^{i}(x_i,\hat{w}^{i}),y_i] \Big\} ,
	\end{align*}
	and then we give the following theorem to illustrate the equivalence of Ro stability and generalization:
	\begin{thm}
		$A(w)$ have Ro stability with rate $ \epsilon_{n} $ if and noly if  
		$A(w)$ generalizes with rate $ \epsilon_{n} $. 
		\label{thm:3-2}
	\end{thm}
	Theorem \ref{thm:3-2} shows that stability is an important property of weight selection criterion, which can ensure that the corresponding estimator has good generalization performance. 
	
	Let $\hat{w}^* \in W$ satisfy $ F(\hat{w}^*,S) = inf_{w \in W} F(w,S) $. Note that 
	\begin{align*}
		& E_{S  } [ F(\hat{w},S) - F(\hat{w}^*,S)   ] \\
		&=  E_{S  } [ F(\hat{w},S) - \hat{F}(\hat{w},S)+ \hat{F}(\hat{w},S) - \hat{F}(\hat{w}^*,S) + \hat{F}(\hat{w}^*,S)  - F(\hat{w}^*,S)  ]   \\
		&\leq E_{S  } [ F(\hat{w},S) - \hat{F}(\hat{w},S)+ \hat{F}(\hat{w},S) - inf_{w \in W} \hat{F}(w,S) + \hat{F}(\hat{w}^*,S)  - F(\hat{w}^*,S)  ] ,
	\end{align*}
	so we give the following theorem to illustrate the relationship between stability and consistency:
	\begin{thm}
		If $A(w)$ is an AERM and has Ro stability with rate $ \epsilon_{n} $, and $\hat{w}^* $ satisfies
		$$ E_{S  } \big[ \hat{F}(\hat{w}^*,S) - F(\hat{w}^*,S)  \big] =O( \epsilon_{n} ) ,$$
		then $A(w)$ is consistent with rate $\epsilon_{n}$.
		\label{thm:3-3}
	\end{thm}
	Since $\hat{w}^*$ and $\mathcal{H}$ depend on $S$, unlike Lemma 15 in \cite{SLT0}, Theorem $\ref{thm:3-3}$ requires $\hat{w}^*$ to generalize with rate $\epsilon_{n}$. In the next section, we will propose a $L_2$-penalty model averaging method and prove that it has stability and consistency under certain reasonable conditions.

\section{$L_2$-penalty Model Averaging }
\label{sec3}

	In the most of existing literatures on model averaging, the theoretical 
	properties are explored under the weight set $W^0 = \{w \in [0,1]^M: 
	\sum_{m=1}^M w_m =1 \} $. From Definition $\ref{def1}$, it is seen that, even if the corresponding 
	weight selection criterion is consistent, such a consistency holds only under the 
	subspace of $R^M$. Therefore, a natural question is whether it is possible to 
	make the weight space unrestricted. What will happen when we do so? We note that unrestricted Granger-Ramanathan method 
	obtains the estimator of the weight vector under $R^M$ by minimizing the 
	sum of squared forecast errors from the combination forecast, but its poor performance is observed when it is compared with some other methods\citep[see][]{DS2008}. On the other hand, in the prediction task, we are more concerned about whether the resulting estimator can predict better, and the estimator that minimizes MSPE in the full space will most likely outperform the estimator that minimizes MSPE in the subspace. Therefore, it is necessary to further develop new research ideas.
	
	\subsection{Model Framework and Estimators}
	\label{sec3-1}
	
	We assume that the response variable $y_i$ and covariates $x_i = (x_{1i},x_{2i},...)$ satisfy the following data generating process:
	\begin{align*}
		y_i=\mu_i+e_i=\sum_{k=1}^{\infty} x_{ki}\theta_{k}+e_i,E(e_i|x_i)=0,E(e_i^2|x_i)=\sigma^2_i,
	\end{align*}
	and M approximating models are given by
	\begin{align*}
		y_i=\sum_{k=1}^{k_m} 
		x_{m(k)i}\theta_{m,(k)}+b_{mi}+e_i,m=1,...,M,
	\end{align*}
	where $b_{mi}= \mu_i -\sum_{k=1}^{k_m} x_{m(k)i}\theta_{m,(k)} $ is the 
	approximating error of the $m$-th approximating model. We assume that the $M$-th approximating model contains all the considered covariates. To simplify notation, we let $x_i = (x_{(1)i},...,x_{(k_M)i})$ throughout the rest of this article, where $x_{(k)i} = x_{M(k)i}$.
	
	Let 
	$y=(y_1,...,y_n)^{'}$, $\mu=(\mu_1,...,\mu_n)^{'}$, $e=(e_1,...,e_n)^{'}$, 
	$b_m=(b_{m1},...,b_{mn})^{'}$. Then, the corresponding matrix form of the 
	true model is $y=X_m\theta_m+b_m+e$, where $\theta_m= 
	(\theta_{m,(1)},...,\theta_{m,(k_m)} )^{'}$, and $X_m$ is the design matrix 
	of the $m$-th approximating model. When there is an approximating model 
	such that $b_m= \textbf{0}$, it indicates that the true model is included 
	in the $m$-th approximating model, i.e. the model is correctly specified. 
	Unlike \cite{HMMA} , \cite{WMMA} and \cite{JMA2012}, we do not require that 
	the infimum of $R_n(w)$ (this is defined in section 
	$\ref{sec3-2}$) tends to infinity, and therefore allow that the model is correctly 
	specified.
	
	Let $\pi_m \in R^{K \times k_m} $ be the variable selection matrix 
	satisfying $ X_M \pi _m =X_m$ and $\pi _m^{'}\pi _m = I_{k_m}$, 
	$m=1,...,M$. Then, the hypothesis spaces of M approximating models are
	$$ \mathcal{H}_m = \big\{ x^{*'} \pi_m \theta_m ,\theta_m \in R^{k_m} \big\},m=1,...,M. $$
	The least squares estimator of $\theta_m$ is
	$\hat{\theta}_m = (X_m^{'} X_m)^{-1} X_m^{'}y$, $m=1,...,M$.

\subsection{Weight Selection Criterion}
\label{sec3-2}

	Let $P_m=X_m (X_m^{'}X_m)^{-1} X_m^{'}$, $P(w) = 
	\sum_{m=1}^M w_m P_m$, $ L_n(w) =\|\mu - P(w)y \|_2^2 $, and $R_n(w) = E_{e}[L_n(w)]$. When 
	$\sigma_i^2 \equiv \sigma^2$, \cite{HMMA} and \cite{WMMA}  used Mallows 
	criterion $ C_n(w) =\|y - \hat{\Omega} w \|_2^2 + 2\sigma^2 w^{'}\kappa$ to 
	select a model weight vector from hypothesis space $\mathcal{H}^0 = 
	\big\{  x^{*'}  \hat{\theta}(w) ,w \in W^0 \big\} $ and 
	proved that the estimator of weight vector asymptotically minimizes $L_n(w)$, 
	where $\hat{\Omega}= (P_1y,...,P_My)$, $\kappa=(k_1,...,k_M)^{'}$, and $\hat{\theta}(w)=\sum_{m=1}^M w_m \pi_m \hat{\theta}_m $. 
	\cite{JMA2012} used Jackknife criterion $ J_n(w) =\|y-\bar{\Omega} w 
	\|_2^2$ to select a model weight vector from hypothesis space 
	$\mathcal{H}^0 $ and proved that the estimator of weight vector  asymptotically 
	minimizes $L_n(w)$ and  $R_n(w)$, where $\bar{\Omega} = \big[y- 
	D_1(I-P_1)y,...,y- D_M(I-P_M)y\big]$ with $D_m =diag[(1-h_{ii}^m)^{-1}]$ and 
	$h_{ii}^m=x_i^{'} \pi_m(X_m^{'}X_m)^{-1} \pi_m^{'}  x_i,i=1,...,n$. 
	
	Different from \cite{HMMA}, \cite{WMMA} and \cite{JMA2012}, we focus on whether the model averaging estimator can asymptotically minimize MSPE when the model weights are not restricted. Let $\hat{\gamma} =(x^{*'}\pi_1\hat{\theta}_1,...,x^{*'}\pi_M\hat{\theta}_M ) $. Then, the risk function and the empirical risk function are defined as: 
	$$ F(w,S) =  E_{z^*  } [y^* -  x^{*'} \hat{\theta}(w) ]^2 =E_{z^*  } (y^* -   \hat{\gamma} w )^2  $$
	and
	$$ \hat{F}(w,S) =  \frac{1}{n} \sum_{i=1 }^n [y_i -  x_i^{'} \hat{\theta}(w) ]^2 =\frac{1}{n}\|y- \hat{\Omega}w\|_2^2  $$
	respectively. Since \cite{HMMA}, \cite{WMMA} and 
	\cite{JMA2012} restrict $w \in W^0$, the corresponding estimators of weight vector  do not necessarily asymptotically minimize $F(w,S)$ on $R^M$. An intuitive way that enables the estimator of weight vector to asymptotically minimize $F(w,S)$ on $R^M$ is to remove the restriction $w \in W^0$ directly. 
	
	Let $\hat{P}$ and $\bar{P}$ be the orthogonal matrices satisfying $\hat{P}^{'} \hat{\Omega}^{'} \hat{\Omega} \hat{P} = diag(\hat{\zeta}_1 ,..., \hat{\zeta}_M )$, $\bar{P}^{'} \bar{\Omega}^{'} \bar{\Omega} \bar{P} = diag(\bar{\zeta}_1 ,..., \bar{\zeta}_M )$, where $\hat{\zeta}_1 \leq ...\leq \hat{\zeta}_M$ and $\bar{\zeta}_1 \leq ...\leq \bar{\zeta}_M$ are  the eigenvalues of $\hat{\Omega}^{'} \hat{\Omega}$ and $\bar{\Omega}^{'} \bar{\Omega}$ respectively. We assume that $E_{z^*}(\hat{\gamma}^{'} \hat{\gamma}  )$, $\hat{\Omega}^{'} \hat{\Omega}$ and $\bar{\Omega}^{'} \bar{\Omega}$ are invertible( this is reasonable under Assumption \ref{ass:3} ), then  
	$$\hat{w}^0 =argmin_{w \in R^M} C_n(w) =(\hat{\Omega}^{'} \hat{\Omega})^{-1}(\hat{\Omega}^{'}y-\sigma^2 \kappa  )  ,$$ 
	$$\bar{w}^0 =argmin_{w \in R^M} J_n(w) =(\bar{\Omega}^{'} \bar{\Omega})^{-1}\bar{\Omega}^{'}y,  $$
	$$\tilde{w} =argmin_{w \in R^M} \hat{F}(w,S) =(\hat{\Omega}^{'} \hat{\Omega})^{-1}\hat{\Omega}^{'}y  ,$$
	and 
	$$\hat{w}^* =argmin_{w \in R^M} F(w,S) =[E_{z^*}(\hat{\gamma}^{'} \hat{\gamma}  )]^{-1}E_{z^*}(\hat{\gamma}^{'} y^*) .$$
	From this, we can see that, in order to satisfy the consistency, 
	$\hat{w}^0$ and $\bar{w}^0$ should be good estimators of $\hat{w}^*$. 
	However, when approximating models are highly correlated, the minimum 
	eigenvalues of $\hat{\Omega}^{'} \hat{\Omega}$ and $\bar{\Omega}^{'} 
	\bar{\Omega}$ may be small so that $\|\hat{w}^0\|_2^2 = \sum_{m=1}^M 
	\frac{a_m^2}{ \hat{\zeta}_m^2} \geq \frac{a_1^2}{ \hat{\zeta}_1^2} $ and 
	$\|\bar{w}^0\|_2^2 = \sum_{m=1}^M \frac{b_m^2}{ \bar{\zeta}_m^2} \geq 
	\frac{b_1^2}{ \bar{\zeta}_1^2}$ are too large, which usually result in 
	extreme weights, where $(a_1,a_2,...,a_M)^{'} =\hat{P}^{'} \hat{\Omega} y 
	$, $(b_1,b_2,...,b_M)^{'} =\bar{P}^{'} \bar{\Omega} y $. Therefore, similar 
	to ridge regression in \cite{RLSEl1970}, we make the following correction 
	to $C_n(w)$ and $J_n(w)$:
	$$C(w,S) =C_n(w) + \lambda_n w^{'} w ,$$
	$$J(w,S) =J_n(w) + \lambda_n w^{'} w,$$ 
	where $\lambda_n >0 $ is a tuning parameter. The above corrections are 
	actually $L_2$-penalty for weight vector. Let $\hat{Z} = (\hat{\Omega}^{'} 
	\hat{\Omega} + \lambda_n I)^{-1}\hat{\Omega}^{'} \hat{\Omega} $ , $\bar{Z}= 
	(\bar{\Omega}^{'} \bar{\Omega} + \lambda_n I)^{-1}\bar{\Omega}^{'} 
	\bar{\Omega}$. Then
	$$\hat{w} =argmin_{w \in R^M} C(w,S) =(\hat{\Omega}^{'} \hat{\Omega} + \lambda_n I)^{-1}(\hat{\Omega}^{'}y-\sigma^2 \kappa  ) = \hat{Z} \hat{w}^0 ,$$ 
	and
	$$\bar{w} =argmin_{w \in R^M} J(w,S) =(\bar{\Omega}^{'} \bar{\Omega} + \lambda_n I)^{-1}\bar{\Omega}^{'}y =\bar{Z} \bar{w}^0  .$$
	In the next subsection, we discuss the theoretical properties of $C(w,S)$ and $J(w,S)$.

\subsection{Stability and Consistency}
\label{sec3-3}

	Let $\lambda_{max}(\cdot)$ and $\lambda_{min}(\cdot)$ be the maximum and 
	minimum eigenvalues of a square matrix respectively, and $\chi$ be the value space of K 
	covariates, where $K=k_M$. In order to discuss the stability and 
	consistency 
	of the proposed method, we need the following assumptions:
	\begin{assumption}
		There are constants $C_1>0$ and $C_2>0$ such that 
		$$C_1 \leq \lambda_{min}(n^{-1} X_{M}^{'}X_{M}) \leq \lambda_{max}(n^{-1} X_{M}^{'}X_{M}) \leq C_2 K , a.s.. $$
		\label{ass:1}
	\end{assumption}
	\begin{assumption}
		There is a constant $C_3>0$ such that $E_{y^*  }[(y^{*})^2] \leq 
		C_3$, and $n^{-1} y^{'}y \leq C_3$, a.s.; There is a constant  $C_4> 0$ such that $\chi 
		\subset B({\bf{0}_K},C_4)$, a.s., where $B({\bf{0}_K},C_4)$  is a ball with center ${\bf{0}_K}$ and radius $C_4$, and ${\bf{0}_K}$ is the K-dimensional 0 vector.
		\label{ass:2}
	\end{assumption}
	\begin{assumption}
		There are constants $C_5>0$ and $C_6>0$ such that 
		$$  C_5 \leq n^{-1} \lambda_{min}( \hat{\Omega}^{'} \hat{\Omega}) \leq n^{-1} \lambda_{max}( \hat{\Omega}^{'} \hat{\Omega}) \leq C_6M , $$
		$$C_5 \leq \lambda_{min}[E_{z^* } (\hat{\gamma}^{'} \hat{\gamma})  ] 
		\leq \lambda_{max}[E_{z^* } (\hat{\gamma}^{'} \hat{\gamma})  ] \leq 
		C_6M ,  $$
		and
		$$C_5 \leq n^{-1} \lambda_{min}( \bar{\Omega}^{'} \bar{\Omega}) \leq  n^{-1} \lambda_{max}( \bar{\Omega}^{'} \bar{\Omega}) \leq C_6M  , $$
		a.s..
		\label{ass:3}
	\end{assumption}
	\begin{assumption}
		$\lambda_n=O(M\log n)$, $\lambda_n - \lambda_{n-1}  = O( K^3M )$.
		\label{ass:4}
	\end{assumption}
	Assumption \ref{ass:1} is mild and similar conditions can be found in \cite{A12009} and \cite{RMMA}. From $ 
	X_{m}^{'}X_{m}= \pi_m^{'} X_{M}^{'}X_{M} \pi_m$, we see that, under 
	Assumption \ref{ass:1}, for any m $\in \{1,...,M\}$, we have
	$$ C_1 \leq \lambda_{min}(n^{-1} X_{m}^{'}X_{m}) \leq \lambda_{max}(n^{-1} X_{m}^{'}X_{m}) \leq C_2 K, a.s..  $$
	\cite{SLT0} assumed that the loss function is bounded, which 
	is usually not satisfied in traditional regression analysis. We replace this assumption 
	with Assumption $\ref{ass:2}$. \cite{MLS} assumed that $\chi \times \mathcal{Y}$ is a 
	compact subset of $R^{K+1}$, under which Assumption \ref{ass:2} is 
	obviously true. Assumption \ref{ass:3} requires the minimum eigenvalue of 
	$\hat{\Omega}^{'} \hat{\Omega}$ to have a lower bound away from 0 and the maximum 
	eigenvalue to have an order $O(nM)$ a.s.. A similar assumption is used 
	in \cite{CMMA2}. Lemma $\ref{lem:3}$ guarantees the rationality of the 
	assumptions about the eigenvalues of $E_{z^* } (\hat{\gamma}^{'} 
	\hat{\gamma})$ and $ \bar{\Omega}^{'} \bar{\Omega}$. Assumption \ref{ass:4} 
	is a mild assumption for tuning parameter in order to avoid excessive 
	penalty. In Section $\ref{sec3-4}$, we look for tuning parameters only from 
	$[0,M\log n]$.
	
	Let $\hat{V}(\lambda_n) =\|\hat{Z} \hat{w}^0 - \hat{Z} \hat{w}^* \|_2^2 $, $\hat{B}(\lambda_n) =\|\hat{Z} \hat{w}^* -  \hat{w}^* \|_2^2 $, $\bar{V}(\lambda_n) =\|\bar{Z} \bar{w}^0 - \bar{Z} \hat{w}^* \|_2^2 $, and $\bar{B}(\lambda_n) =\|\bar{Z} \hat{w}^* -  \hat{w}^* \|_2^2 $. We define
	$$\hat{M}(\lambda_n) =\|\hat{Z} \hat{w}^0 -  \hat{w}^* \|_2^2 = \hat{V}(\lambda_n)  +\hat{B}(\lambda_n) + 2(\hat{Z} \hat{w}^0 - \hat{Z} \hat{w}^*)^{'} (\hat{Z} \hat{w}^* -  \hat{w}^*)   ,$$
	$$\bar{M}(\lambda_n) =\|\bar{Z} \bar{w}^0 -  \hat{w}^* \|_2^2 = \bar{V}(\lambda_n)  +\bar{B}(\lambda_n) + 2(\bar{Z} \bar{w}^0 - \bar{Z} \hat{w}^*)^{'} (\bar{Z} \hat{w}^* -  \hat{w}^*)   .$$
	In order for $F(\hat{Z} \hat{w}^0,S)$ and $F(\bar{Z} \bar{w}^0,S)$ to better approximate $F( \hat{w}^*,S)$, we naturally want $E_{S   }[\hat{M}(\lambda_n) ]$ and $E_{S   }[\bar{M}(\lambda_n) ]$ to be as small as possible. In the following discussion, we call $E_{S   }[\hat{M}(\lambda_n)] $ and $E_{S   }[\bar{M}(\lambda_n)] $, $E_{S   }[\hat{V}(\lambda_n)]$ and $E_{S   }[\bar{V}(\lambda_n)]$, $E_{S   }[\hat{B}(\lambda_n)]$ and $E_{S   }[\bar{B}(\lambda_n)]$ the corresponding mean square errors, estimation variances, and estimation biases, respectively. Obviously, when $\lambda_n=0$, $\hat{Z} = \bar{Z} = I_{M}$ which means estimation bias is equal to zero. From Lemma $\ref{lem:4}$ and the proof of Theorems 
	$\ref{thm:4-2-1}$, we see that, under Assumptions $\ref{ass:1}-\ref{ass:4}$,  $\hat{B}(\lambda_n)$ and 
	$\bar{B}(\lambda_n)$ are $O( n^{-2}M^4log^2n)$, a.s.. On the other hand, the existence of extreme weights may make the performance of $\hat{w}^0$ and  $\bar{w}^0$ extremely unstable. So the purpose of using $L_2$-penalty is to reduce estimation variance by introducing estimation bias, and thus make the performance of the model average estimator more stable. Further we define
	$$\hat{M}_1(\lambda_n) = \hat{V}(\lambda_n)  +\hat{B}(\lambda_n)    ,$$
	$$\bar{M}_1(\lambda_n) =  \bar{V}(\lambda_n)  +\bar{B}(\lambda_n)    .$$
	Like Theorem 4.3 in \cite{RLSEl1970}, we give the following theorem to illustrate the reasonability of introducing $L_2$-penalty:
	\begin{thm}
		Let $\hat{\lambda}_n= \min \{\lambda_n: \frac{d}{d \lambda_n } \hat{M}_1(\lambda_n) =0 \} $, $\bar{\lambda}_n = \min \{\lambda_n: \frac{d}{d \lambda_n } \bar{M}_1(\lambda_n) =0 \}$. Then, 1) when $\hat{w}^0 \neq  \hat{w}^*$, $\hat{\lambda}_n >0$ and $ \hat{M}_1(\hat{\lambda}_n) < \hat{M}_1(0)$; 2) when $\bar{w}^0 \neq  \hat{w}^*$, $\bar{\lambda}_n>0$ and $\bar{M}_1(\bar{\lambda}_n) < \bar{M}_1(0)$.
		\label{thm:4-2-1}
	\end{thm}
	Theorems \ref{thm:4-2-1} shows that the use of 
	$L_2$-penalty reduces estimation variance by introducing estimation bias. 
	However, since $\hat{w}^*$ is unknown, $\hat{\lambda}_n$ and 
	$\bar{\lambda}_n$ are also unknown. In Section $\ref{sec3-4}$, we use cross 
	validation to select the tuning parameter $\lambda_n$. The following theorem 
	shows that $C(w,S)$ and $J(w,S)$ are AERM.
	\begin{thm}
		Under Assumptions $\ref{ass:1}-\ref{ass:4}$, $C(w,S)$ and $J(w,S)$ are AERM with rate $n^{-1} \log n K M^2 ( 1+n^{-2}K^2)$ and $n^{-1} \log n K^3  M^2$, respectively.
		\label{thm:4-3-1}
	\end{thm}
	The following theorem shows that $C(w,S)$, $J(w,S)$ and $F(w,S)$ have FLoo stability and PLoo stability.
	\begin{thm}
		Under Assumptions $\ref{ass:1}-\ref{ass:4}$, $C(w,S)$, $J(w,S)$ and $F(w,S)$ have FLoo stability and 
		PLoo stability with rate $n^{-\frac{1}{2}}  K^{\frac{7}{2}} 
		M^{\frac{5}{2}} ( 1+n^{-2}K^2)$, $n^{-\frac{1}{2}} K^{\frac{7}{2}} 
		M^{\frac{5}{2}} $ and $n^{-\frac{1}{2}} K^{\frac{7}{2}} M^{\frac{7}{2}} $
		, respectively.
		\label{thm:4-3-2}
	\end{thm}
	It can be seen from Theorems $\ref{thm:3-1}$, $\ref{thm:3-2}$ and $\ref{thm:4-3-2}$ that $C(w,S)$, $J(w,S)$ and $F(w,S)$ have Ro stability and generalization. The following theorem shows that $C(w,S)$ and $J(w,S)$ have consistency, which is a direct consequence of Theorems $\ref{thm:3-1}-\ref{thm:3-3}$ and $\ref{thm:4-3-1}-\ref{thm:4-3-2}$.
	\begin{thm}
		Under Assumptions $\ref{ass:1}-\ref{ass:4}$, $C(w,S)$ and $J(w,S)$ have consistency with rate 
		$n^{-\frac{1}{2}} K^{\frac{7}{2}} M^{\frac{7}{2}} ( 1+n^{-2}K^2)$ and 
		$n^{-\frac{1}{2}} K^{\frac{7}{2}} M^{\frac{7}{2}} $, respectively.
		\label{thm:4-3-3}
	\end{thm}	

\subsection{ Optimal Weighting Based on Cross Validation  }
\label{sec3-4}

	Although Theorems \ref{thm:4-2-1} shows that there are 
	$\hat{\lambda}_n$ and $\bar{\lambda}_n$  such that $\hat{w}$ and $\bar{w}$ 
	are better approximation of $\hat{w}^*$,  $\hat{\lambda}_n$ and 
	$\bar{\lambda}_n$ cannot be obtained. Therefore, like 
	\cite{SMA}, we propose an algorithm based on 10-fold cross validation to 
	obtain the estimator of weight vector, which is a weighted average of the weight  estimators for different $\lambda_n$. That is, we first select 100 segmentation points on $[0,M\log 
	n]$ with equal intervals as the candidates of $\lambda_n$. Then we calculate the estimation error for each candidate of 
	$\lambda_n$ by using 10-fold cross validation. Based on this, we remove those candidates with large estimation error. Last, for remaining candidates, the estimation errors are used to perform a weighted 
	average of the estimators of weight vector. We summarize our algorithm on RMMA below, and a similar algorithm can be given for RJMA.
	\begin{center}
		\begin{breakablealgorithm}
			\caption{Optimal weighting based on cross validation }
			\label{alg1}
			\begin{algorithmic}[1] 
				\Require S;
				\Ensure $\hat{w}$;
				\State $\hat{E}_L = 0, L=1,...,100$; 
				\State The sample set S is randomly divided into 10 sample subsets with equal size, and the sample index set belonging to the B-th part is 
				denoted as $S_{B},B=1,...,10$;
				
				\For{ each $B \in \{1,2,...,10\}$}
				\State Let $S_{train}= \{ z_i , i \notin S_{B}  \}$ which is the training set, and $S_{test}= \{ z_i , i \in S_{B}  \}$ which is the testing set;
				\State $\hat{\theta}_m^B$ is obtained based on $S_{train}$ , $m=1,...,M$;
				\For{ each $L \in \{1,2,...,100\}$}
				\State $\hat{w}_{BL}$ is obtained based on 
				$\lambda_n=\frac{ (L-1) M\log n}{99}$ and 
				$C(w,S_{train})$;
				\State The estimation error of $\hat{w}_{BL}$ on $S_{test}$ is obtained as
				$$\hat{E}(\hat{w}_{BL}) =  \sum_{z_i \in S_{test}} [y_i - x_i^{'}\hat{\theta}^B(\hat{w}_{BL})] ^2 ,$$
				where $\hat{\theta}^B(w) = \sum_{m=1}^M w_m \pi_m \hat{\theta}_m^B$;
				\State $\hat{E}_L =\hat{E}_L+ \hat{E}(\hat{w}_{BL})$;
				\EndFor
				\EndFor
				\State Let $S_{\lambda}$ be the index set of the smallest 50 
				numbers in $\{\hat{E}_L, L=1,...,100\}$;
				\State $\hat{w}_L$ is obtained based on $\lambda_n=\frac{ (L-1) 
					M\log n}{99}$, $S$, and 
				$C(w,S)$, where $L \in 
				S_{\lambda}$;
				\State $\hat{w}= \sum_{L \in S_{\lambda}} \frac{ exp(-0.5\hat{E}_L) }{\sum_{L \in S_{\lambda} exp(-0.5\hat{E}_L) } } \hat{w}_L $.
			\end{algorithmic}
		\end{breakablealgorithm}
	\end{center}

\section{Simulation Study  }
\label{sec4}
	In this section, we conduct simulation experiments to demonstrate the finite sample performance
	of the proposed method. 
	Similar to \cite{HMMA}, we consider the following data generating process:
	\begin{align*}
		y_i= \mu_i + e_i = \sum _{k=1}^{K_t} x_{ki} \theta _{k} +e _i\ \ i=1,...,n,
	\end{align*}
	where $\{\theta _{k},k=1,2,...,K_t\}$ are the model parameters, $x_{1i} \equiv 1$, $(x_{2i},...,x_{K_ti}) \sim N(0,\Sigma)$, and $(e_1,e_2,...,e_n) \sim N[0, diag(\sigma_1^2,\sigma_2^2,...,\sigma_n^2) ]$. We set $n=100,300,500,700$, $\alpha = 0.5,1.0,1.5$, $\Sigma=(\sigma_{kl})$ and $\sigma_{kl}=\rho^{|k-l|}$ with $\rho=0.3,0.6 $, and $R^2 =  0.1,...,0.9$, where the population $R^2 = \frac{var( \sum _{k=1}^{K_t} x_{ki} \theta _{k} )  }{ var( \sum _{k=1}^{K_t} x_{ki} \theta _{k} + e_i ) }$. For the homoskedastic simulation we set $\sigma_i^2 \equiv 1$, while for the
	heteroskedastic simulation we set $\sigma_i^2 = x_{2i}^2$.
	
	We compare the following ten model selection/averaging methods: 1) model selection with AIC (AI), model selection 
	with BIC (BI), and model selection with $C_p$ (Cp); 2) model averaging with 
	smoothed AIC (SA), and model averaging with smoothed BIC (SB); 3) Mallows model 
	averaging (MM), jackknife model averaging (JM), and least squares model 
	averaging based on generalized cross validation (GM)\citep{GMMA}; 4) 
	Ridge-Mallows model averaging (RM), and Ridge-jackknife model averaging (RJ). 
	To evaluate these methods, we generate test set $ \{ (\mu_i^*, 
	x_i^*),i=1,...,n_t  \} $ by the above data generating process, and
	$$ MSE = n_t^{-1} \sum_{i=1}^{n_t}   [\mu_i^* - x_i^{*'} \hat{\theta}(\hat{w}) ]^2 $$
	is calculated as a measure of consistency. In the simulation, we set $n_t = 1000$ and repeat 200 times. For each parameterization, we normalize the MSE by dividing by the infeasible MSE(the mean of the smallest MSEs of M approximating models in 200 simulations).
	
	We consider two simulation settings. In the first setting, like \cite{HMMA}, all candidate models are misspecified, and candidate models are strictly nested. In the second setting, the true model is one of candidate models, and candidate models are non-nested.

\subsection{Nested Setting and Results}

\begin{figure}[ht]
		\centering
		\includegraphics[scale=0.9]{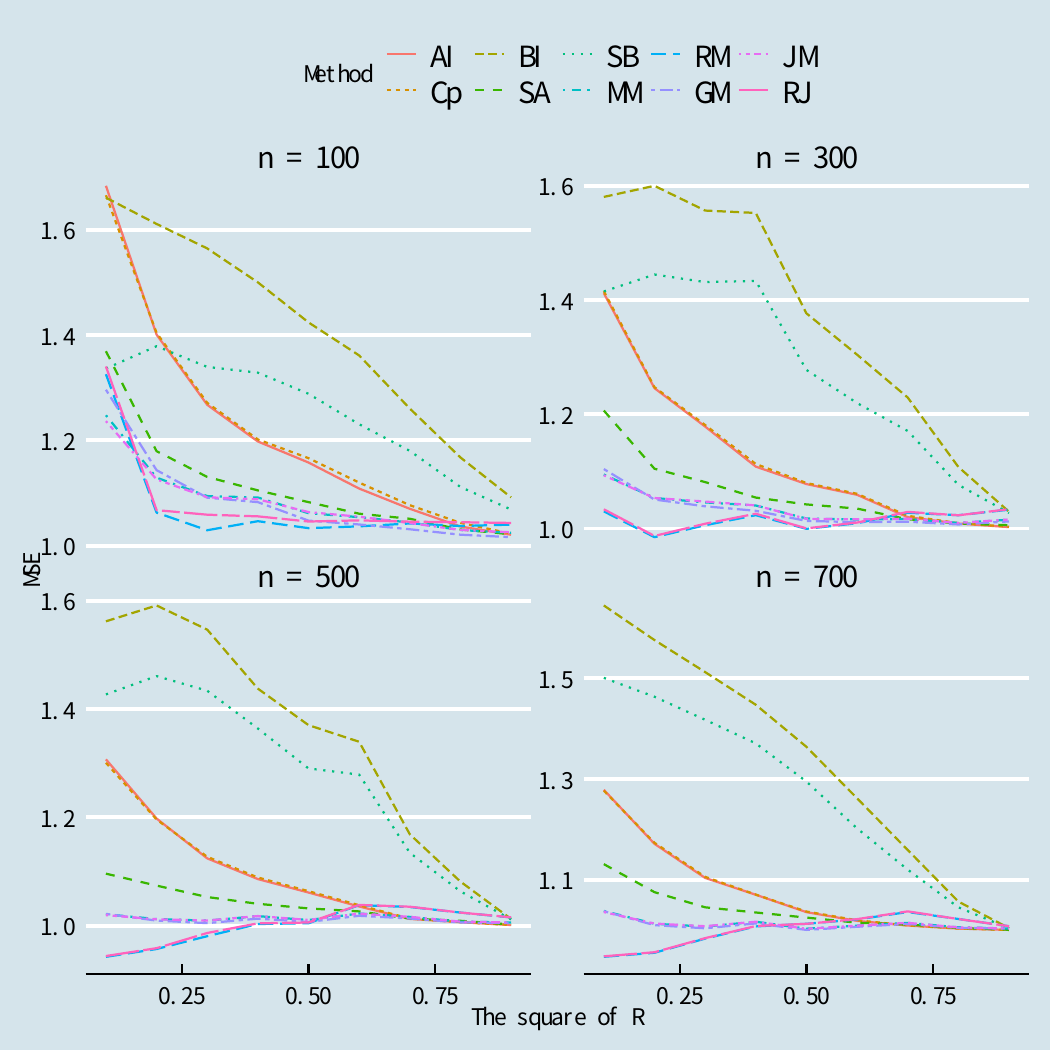}
		\caption{The mean of normalized MSE under homoskedastic errors with $\alpha=0.5$ in nested setting}
		\label{pic:1}
	\end{figure}

\begin{figure}[ht]
		\centering
		\includegraphics[scale=0.9]{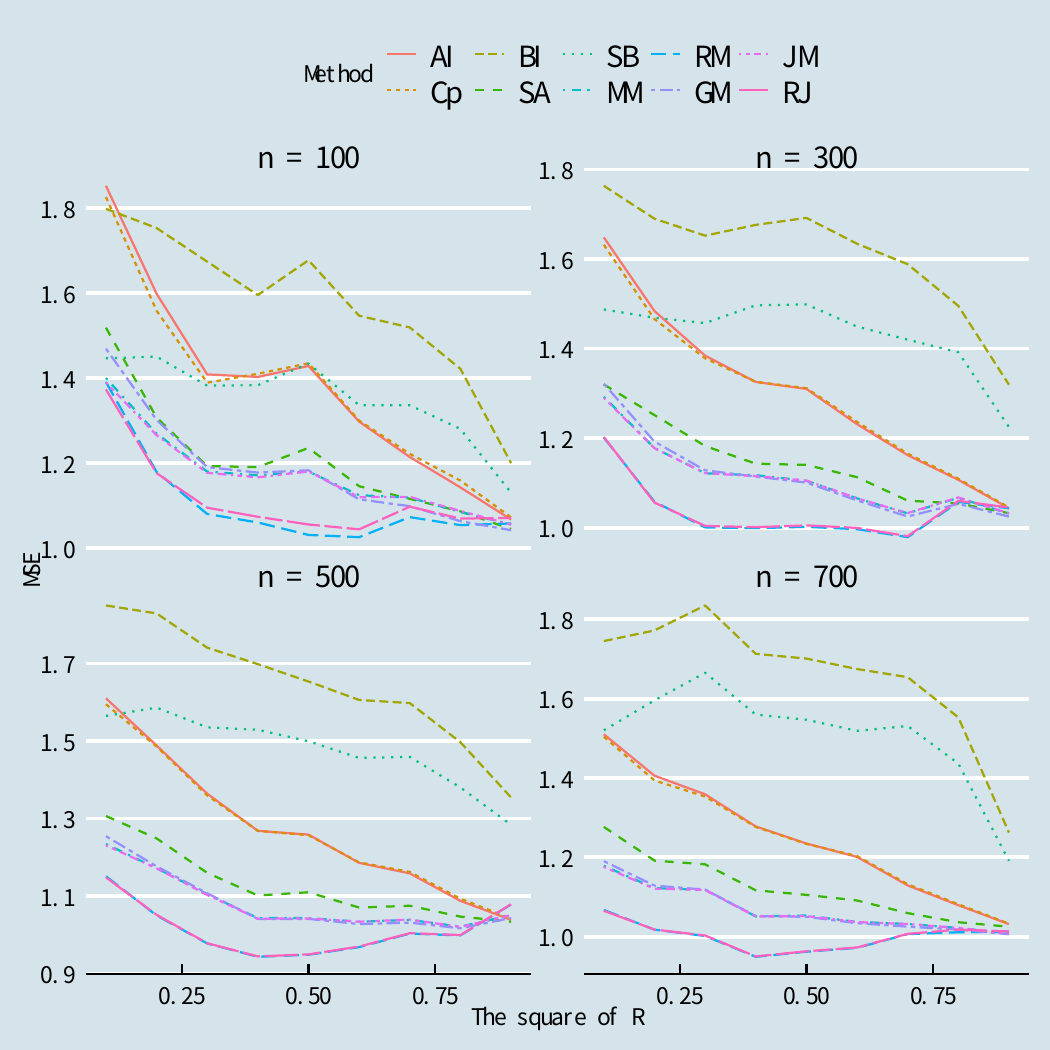}
		\caption{The mean of normalized MSE under homoskedastic errors with $\alpha=1.0$ in nested setting}
		\label{pic:2}
	\end{figure}

\begin{figure}[ht]
		\centering
		\includegraphics[scale=0.9]{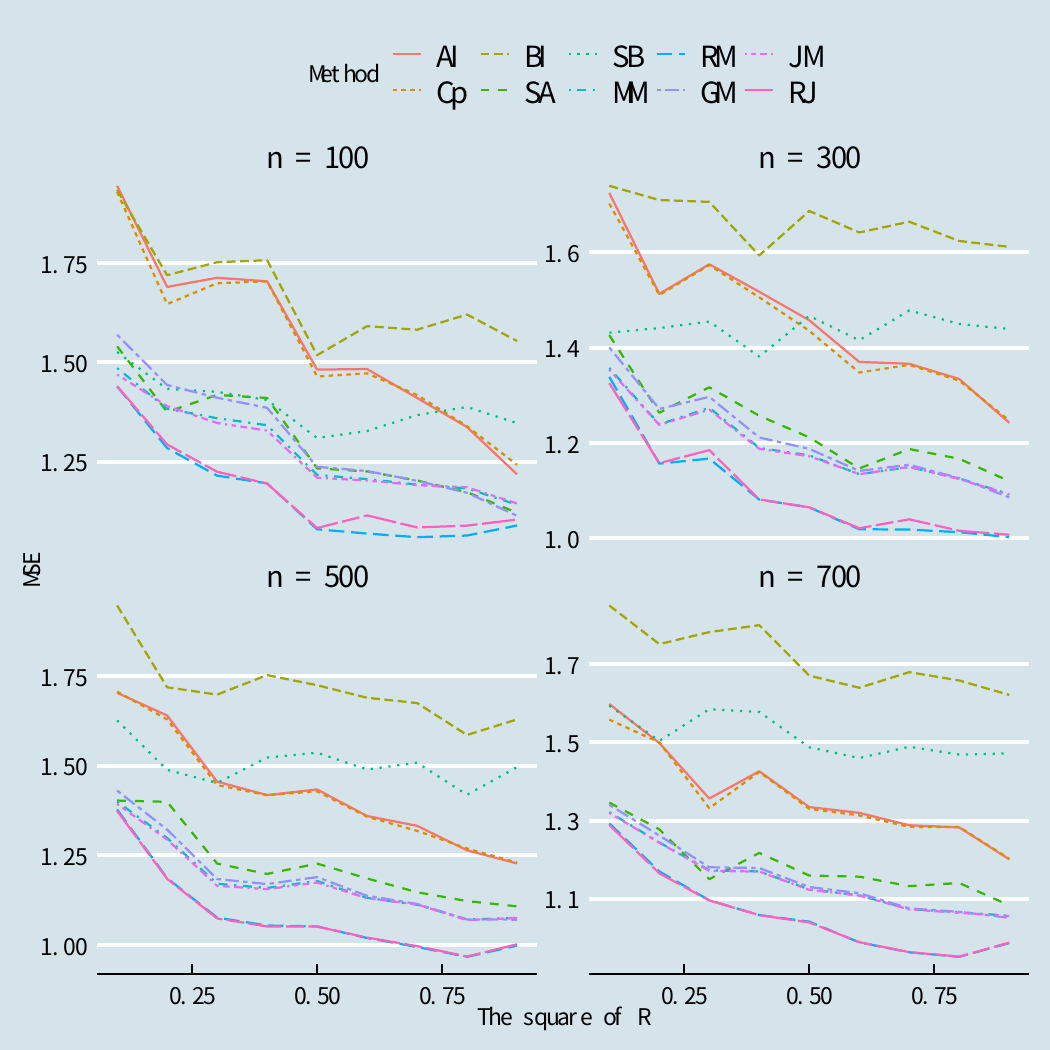}
		\caption{The mean of normalized MSE under homoskedastic errors with $\alpha=1.5$ in nested setting}
		\label{pic:3}
	\end{figure}

\begin{figure}[ht]
		\centering
		\includegraphics[scale=0.9]{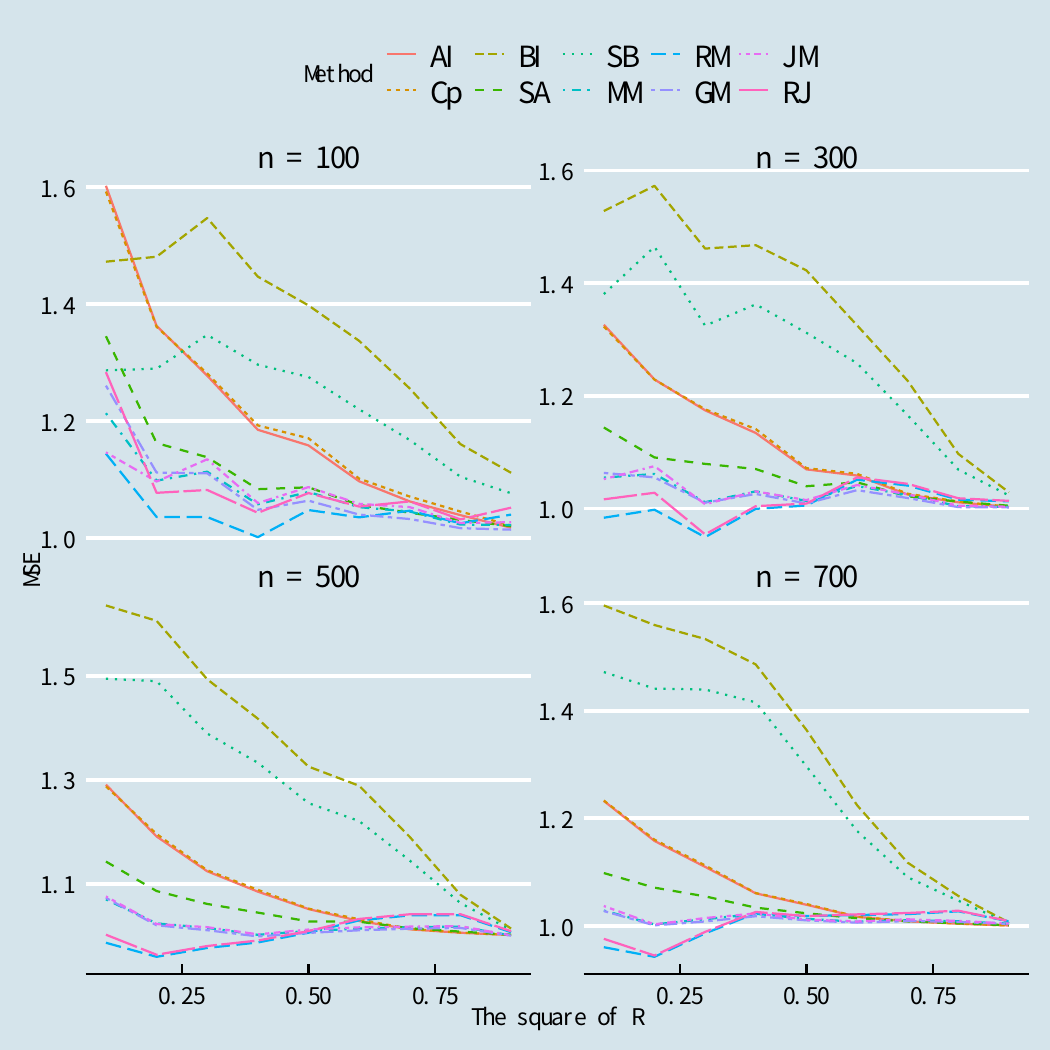}
		\caption{The mean of normalized MSE under heteroskedastic errors with $\alpha=0.5$ in nested setting}
		\label{pic:4}
	\end{figure}

\begin{figure}[ht]
		\centering
		\includegraphics[scale=0.9]{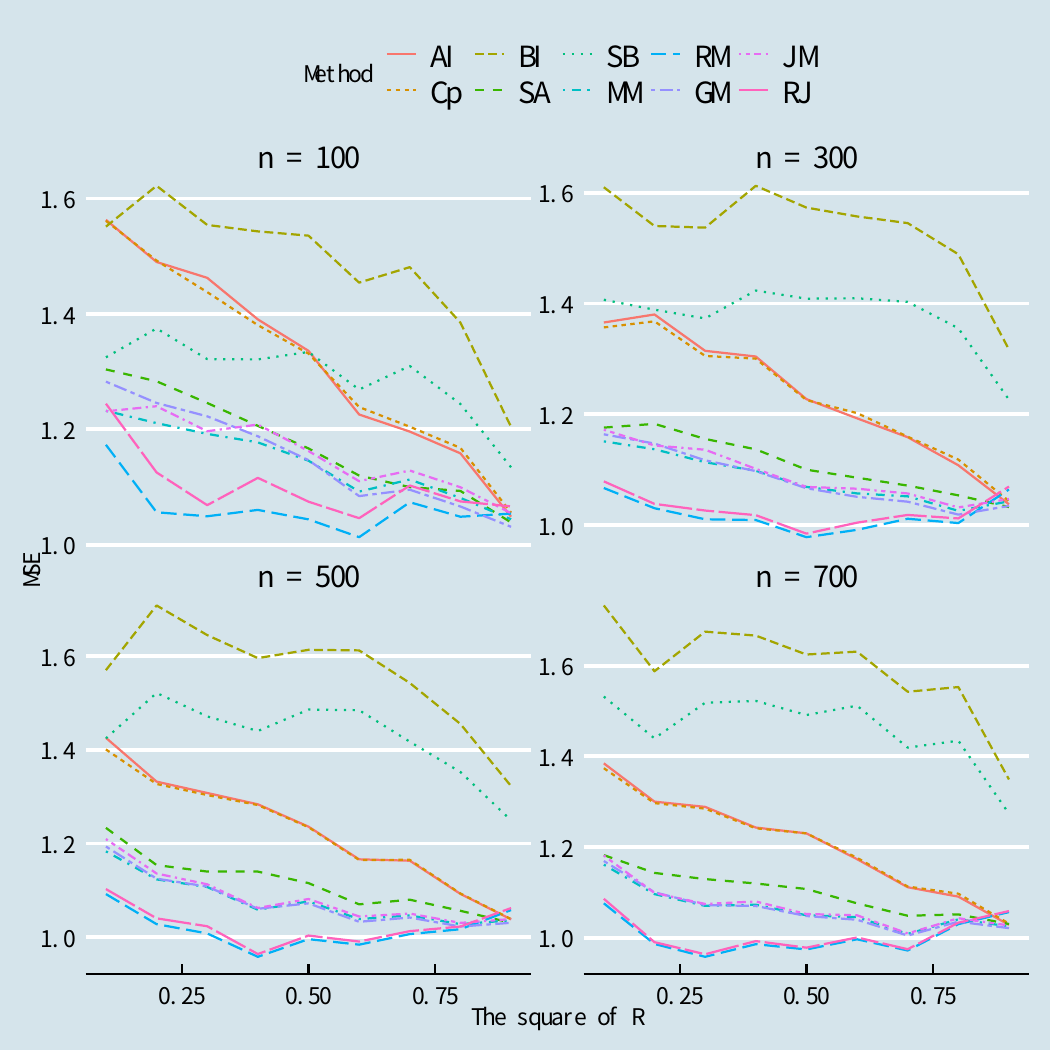}
		\caption{The mean of normalized MSE under heteroskedastic errors with $\alpha=1.0$ in nested setting}
		\label{pic:5}
	\end{figure}

\begin{figure}[ht]
		\centering
		\includegraphics[scale=0.9]{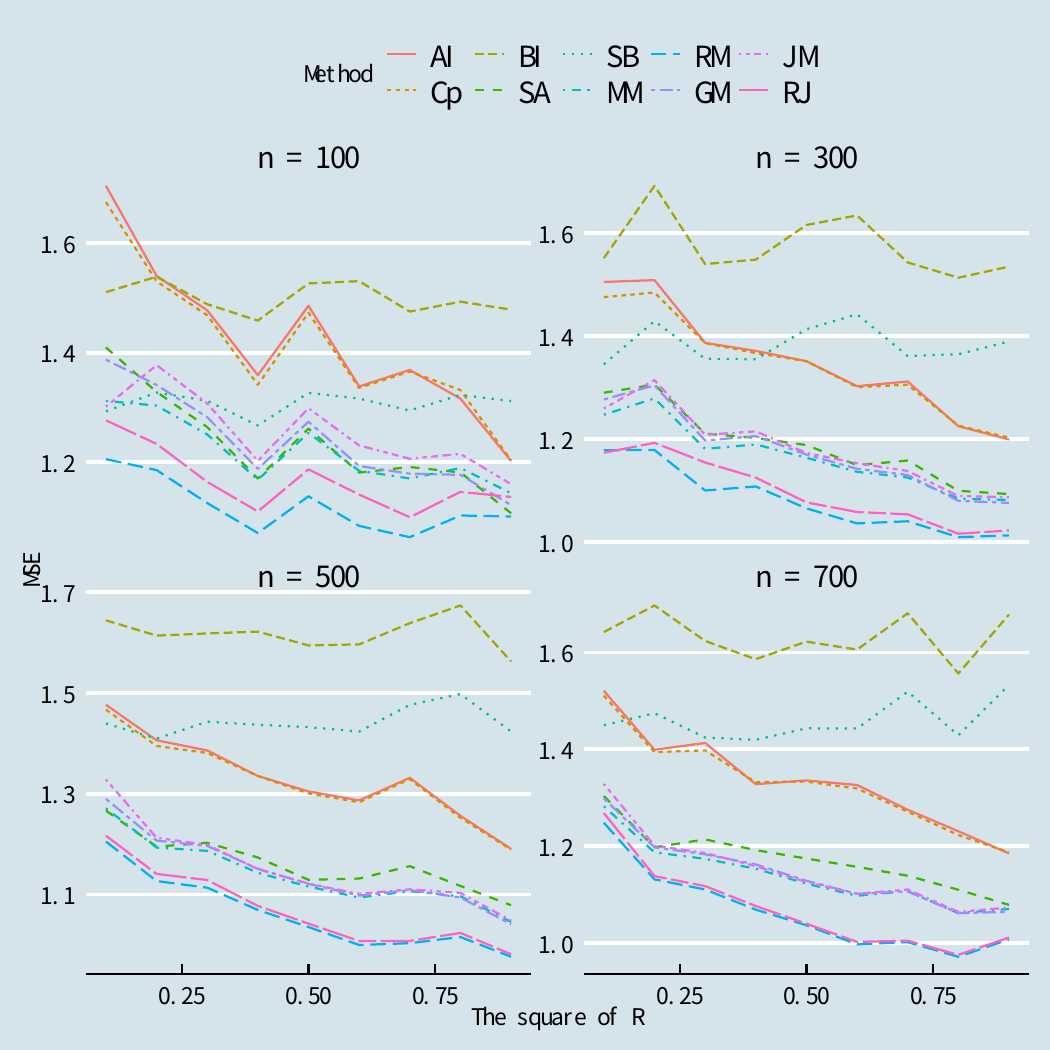}
		\caption{The mean of normalized MSE under heteroskedastic errors with $\alpha=1.5$ in nested setting}
		\label{pic:6}
	\end{figure}

	We set $K_t=400$, $\theta _{k}= c \sqrt{2 \alpha } 
	k^{-\alpha-\frac{1}{2}}$, and $K=\log_{4}^{2}n$(i.e. $K=11,17,20,22$), where $R^2 $ 
	is controlled by the parameter c. For $\rho=0.3$, the mean of normalized MSEs in 200 
	simulations is shown in Figures $\ref{pic:1}-\ref{pic:6}$. The results with $\rho=0.6$ are similar and so omitted for saving space. 
	
	For homoskedastic case, we can draw the following conclusions from Figures 
	$\ref{pic:1}-\ref{pic:3}$. When $\alpha =0.5$ (Figure $\ref{pic:1}$): 1) RM 
	and RJ perform better than other methods if $R^2 \leq 0.5$ and 
	$n=300, 500$ and $700$; 2) RM and RJ perform worse than other methods except for 
	BI and SB if $R^2 > 0.5$, and worse than MA, JM and GM 
	if $R^2 = 0.1$ and $n=100$; 3) RM performs better than RJ in most of 
	cases if $n =100$ and $300$. When $\alpha =1.0$ (Figure $\ref{pic:2}$): 1) RM 
	performs better than other methods if $R^2 \leq 0.8$; 2) RM and RJ 
	perform worse than other methods except for BI and SB if $R^2 = 0.9$; 3) 
	RM performs better than RJ in most of cases if $n =100$ and $300$. When 
	$\alpha 
	=1.5$ (Figure $\ref{pic:3}$): 1) RM and RJ always perform better than other 
	methods; 2) RM performs better than RJ in most of cases if $n=100$ and $300$.
	
	For heteroskedastic case, we can draw the following conclusions from 
	Figures $\ref{pic:4}-\ref{pic:6}$. When $\alpha =0.5$ (Figure 
	$\ref{pic:4}$): 1) RM performs better than other methods if $R^2 \leq 
	0.5$; 2) RM and RJ perform worse than other methods except for BI and SB 
	if $R^2 > 0.5$; 3) RM performs better than RJ. When $\alpha =1.0$ (Figure 
	$\ref{pic:5}$): 1) RM performs better than other methods if $R^2 \leq 
	0.7$; 2) RM and RJ perform worse than other methods except for BI and SB 
	if $R^2 > 0.7$; 3) RM performs better than RJ, and RJ performs better 
	than other methods in most cases. When $\alpha =1.5$ (Figure 
	$\ref{pic:6}$): 1) RM always performs better than other methods; 2) RM 
	always performs better than RJ, and RJ performs better than other methods 
	in most cases. 
	
	From Figures $\ref{pic:1}-\ref{pic:6}$, we can draw the following conclusions: 1) As $\alpha$ increases, RM and RJ perform better and better; 2) $n$ has no obvious influence on the performance comparison of various methods; 3) With the increase of $R^2$, the performance of various methods is improved, but RJ and RM always have not very bad performance, and are the best in most cases.
	
	To sum up, the conclusions are as follows: 1) RM and RJ are the best in 
	most cases and not bad if not the best; 2) When $\alpha$ is small and $R^2$ is large, GM has better performance, and RM and RJ are the best in other cases; 3) RM 
	performs better than RJ.

\subsection{Non-nested Setting and Results}

\begin{figure}[ht]
		\centering
		\includegraphics[scale=0.9]{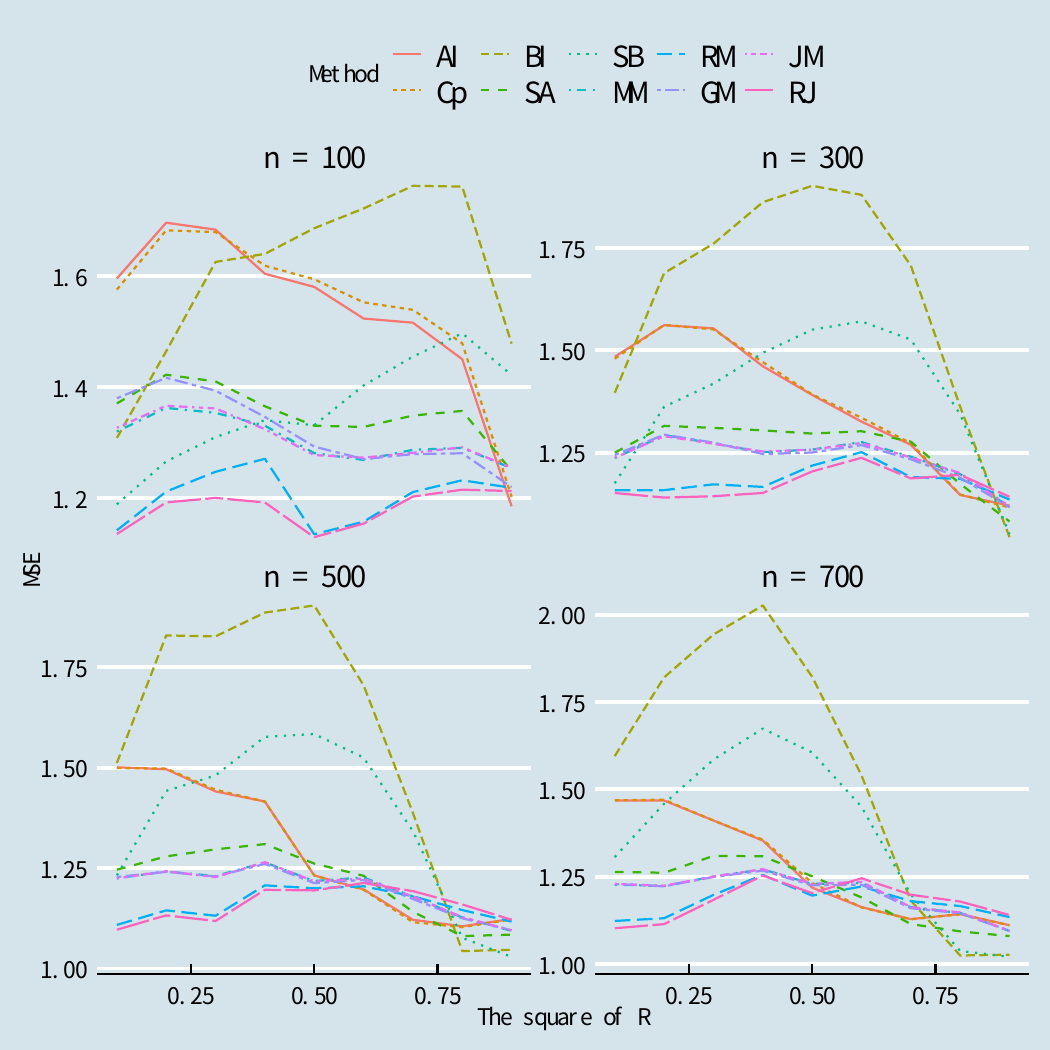}
		\caption{The mean of normalized MSE under homoskedastic errors with $\alpha=0.5$ in non-nested setting}
		\label{pic:7}
	\end{figure}

\begin{figure}[ht]
		\centering
		\includegraphics[scale=0.9]{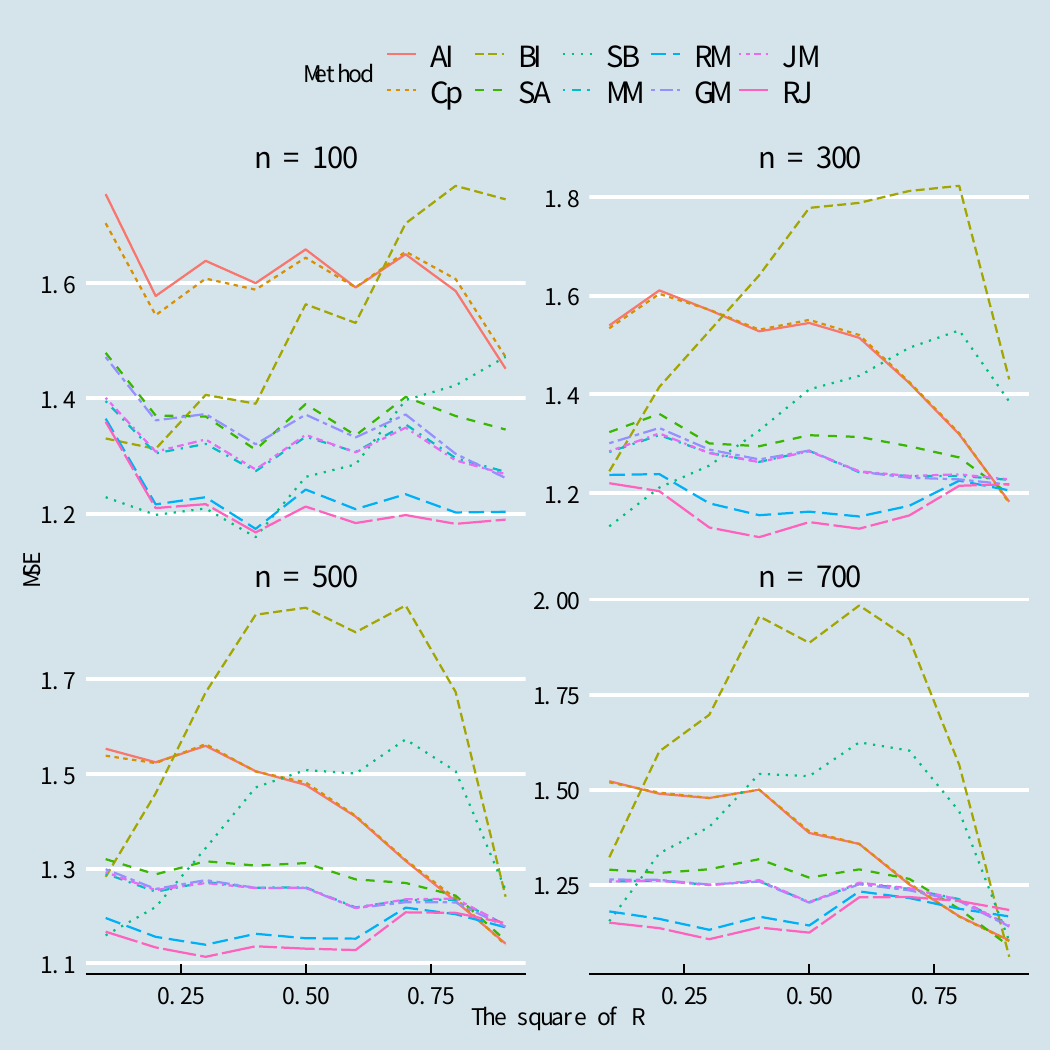}
		\caption{The mean of normalized MSE under homoskedastic errors with $\alpha=1.0$ in non-nested setting}
		\label{pic:8}
	\end{figure}

\begin{figure}[ht]
		\centering
		\includegraphics[scale=0.9]{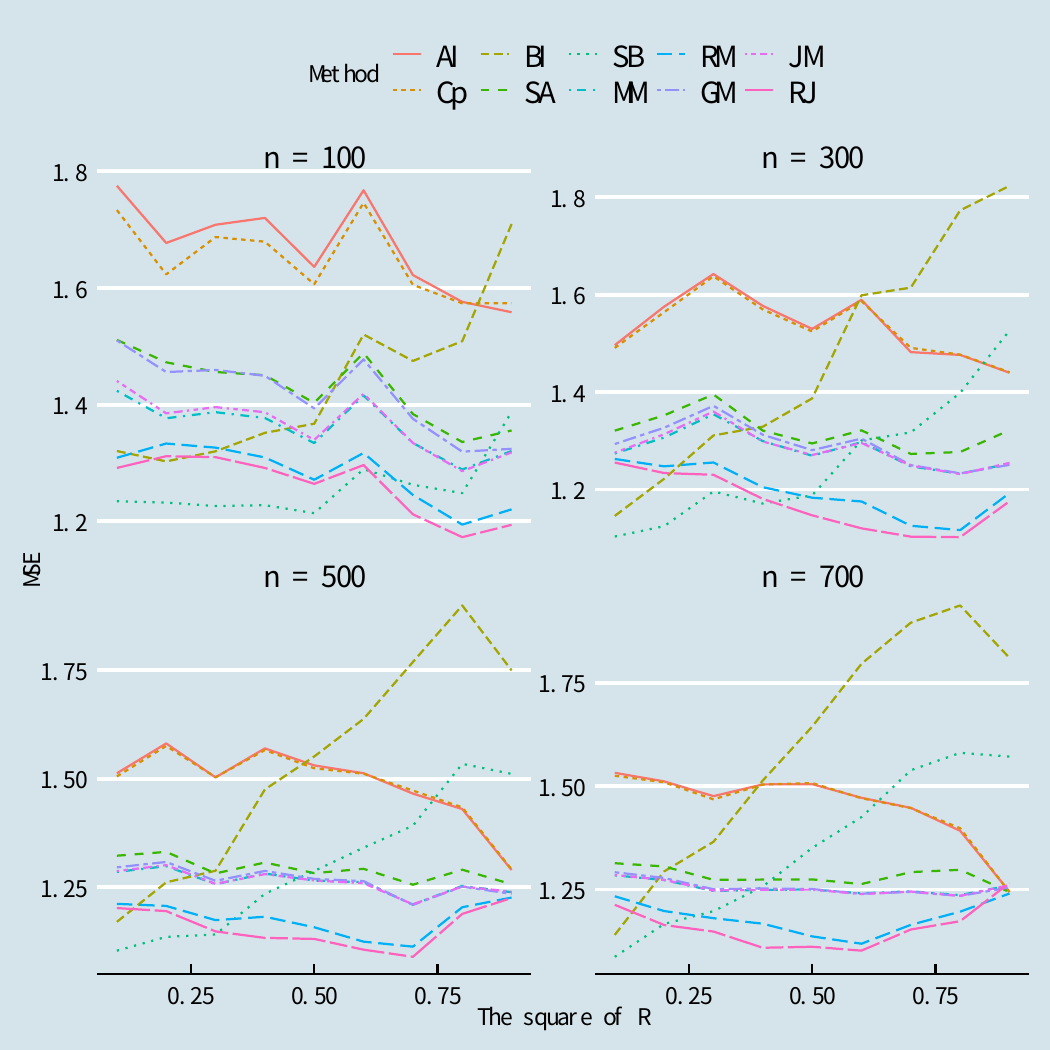}
		\caption{The mean of normalized MSE under homoskedastic errors with $\alpha=1.5$ in non-nested setting}
		\label{pic:9}
	\end{figure}

\begin{figure}[ht]
		\centering
		\includegraphics[scale=0.9]{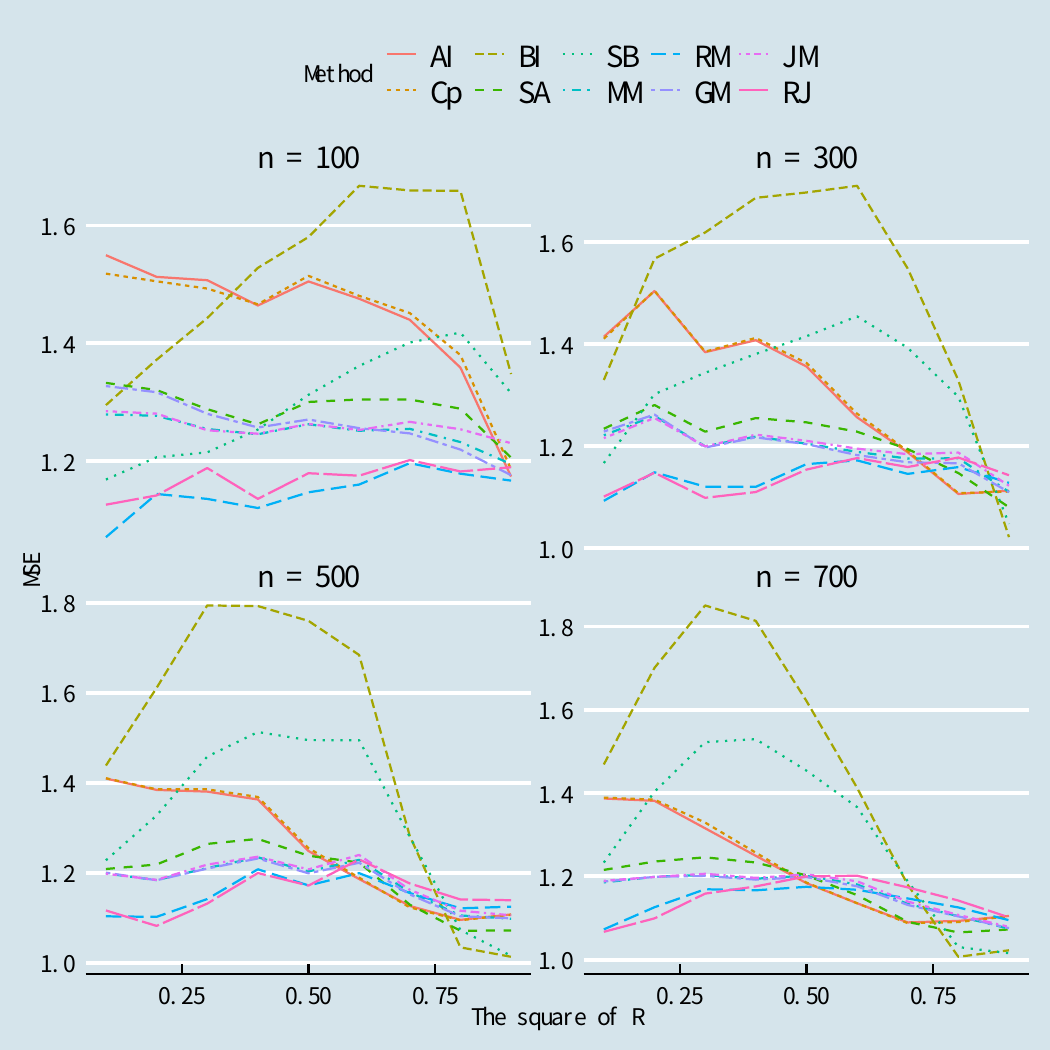}
		\caption{The mean of normalized MSE under heteroskedastic errors with $\alpha=0.5$ in non-nested setting}
		\label{pic:10}
	\end{figure}

\begin{figure}[ht]
		\centering
		\includegraphics[scale=0.9]{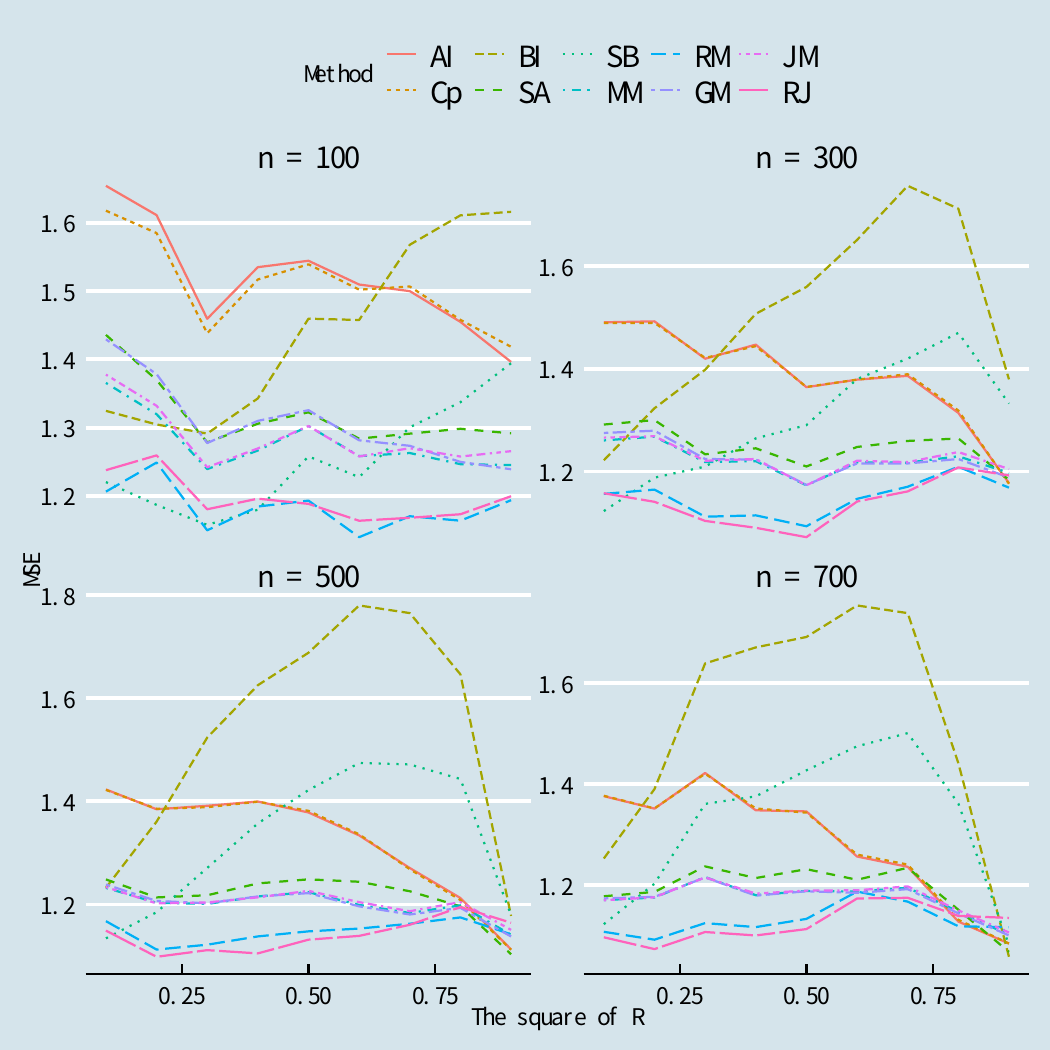}
		\caption{The mean of normalized MSE under heteroskedastic errors with $\alpha=1.0$ in non-nested setting}
		\label{pic:11}
	\end{figure}

\begin{figure}[ht]
		\centering
		\includegraphics[scale=0.9]{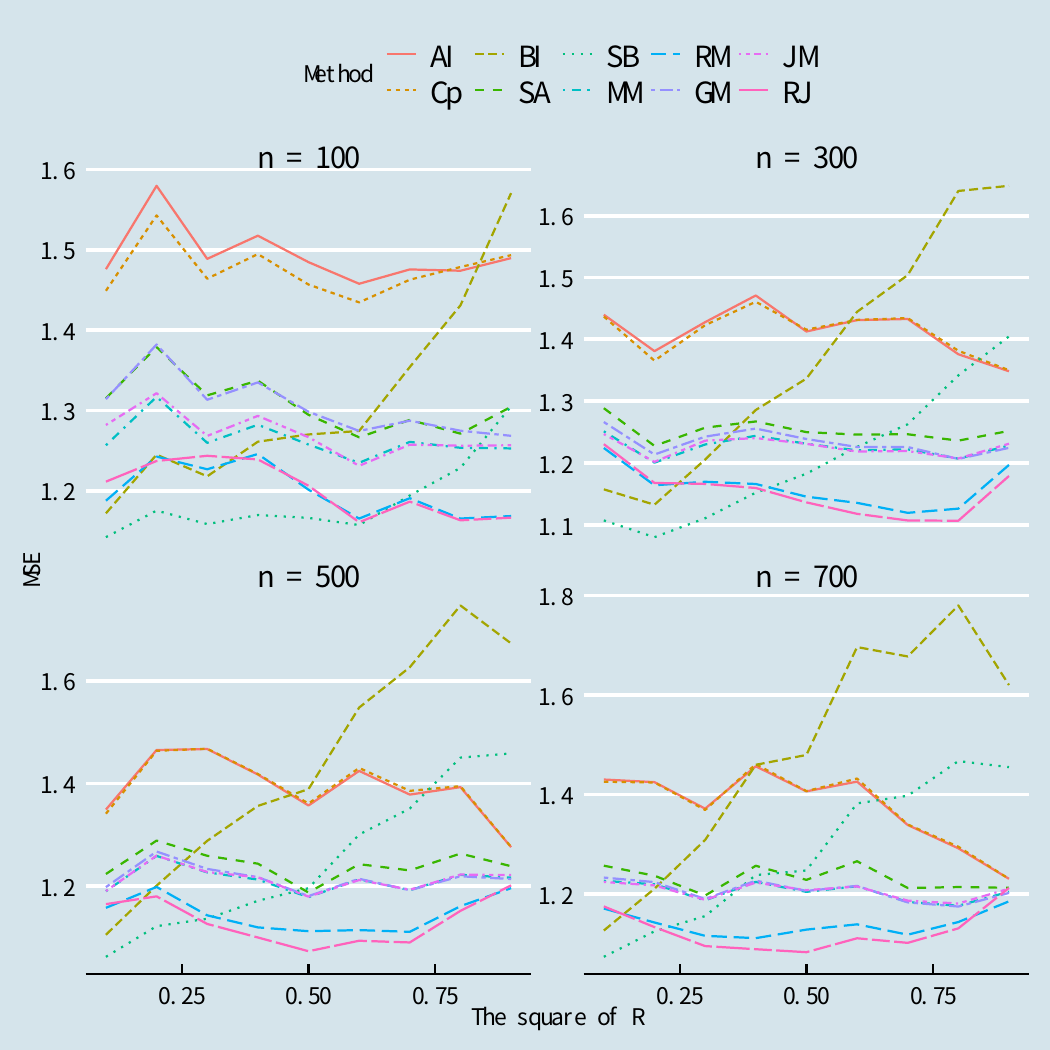}
		\caption{The mean of normalized MSE under heteroskedastic errors with $\alpha=1.5$ in non-nested setting}
		\label{pic:12}
	\end{figure}

	We set $K_t=12$ and $\theta _{k}= c \sqrt{2 \alpha } k^{-\alpha-\frac{1}{2}}$ for $1 \leq k \leq 10$, and $\theta _{k}= 0$ for $k = 11,12$, where $R^2$ is controlled by the parameter c. Each approximating model contains the first 6 covariates, and the last 6 covariates are combined to obtain $2^6$ approximating models. For $\rho=0.3$, the mean of normalized MSEs in 200 simulations is shown in Figures $\ref{pic:7}-\ref{pic:12}$. Like the nested case, the results with $\rho=0.6$ are similar and so omitted.

	For this setting, we can draw the following conclusions from Figures 
	$\ref{pic:7}-\ref{pic:12}$. When $\alpha =0.5$ (Figure $\ref{pic:7}$ and Figure 
	$\ref{pic:10}$): 
	1) RM and RJ perform better than other methods if $R^2 \leq 0.5$; 2) As 
	$n$ increases, RM and RJ perform worse than other methods for larger $R^2$; 
	3) For homoskedastic case, RJ performs better than RM in most of cases. When $\alpha =1.0$ (Figure 
	$\ref{pic:8}$ and Figure $\ref{pic:11}$): 1) RM and RJ perform better than other methods except for 
	SB if $R^2\leq 0.7$, but the performance of SB is very 
	instable; 2) As $n$ increases, RM and RJ perform worse 
	than other methods for larger $R^2$; 3) RJ performs better than RM in most 
	of cases. When $\alpha =1.5$ (Figure $\ref{pic:9}$ and Figure 
	$\ref{pic:12}$): 1) RM and RJ perform 
	better than other methods except for BI and SB, but the performance of SB 
	and BI is very instable; 2) RJ performs better than RM in most of cases.

	To sum up, the conclusions are as follows: 1) RM and RJ are the best in 
	most cases and have stable performance; 2) One of SB, SA, BI, and AI may perform best when R2 is small or large, but their performance is unstable compared to RM and RJ; 3) In heteroskedastic case 
	with larger $\alpha$ and $n$ and homoskedastic case, on the whole, RJ 
	performs better than RM.

\section{Real Data Analysis}
\label{sec5}
	In this section, we apply the proposed method to the real "wage1" dataset in \cite{IE2003} from from
	the US Current Population Survey for the year 1976. There 
	are 526 observations in this dataset. The response variable is the log of 
	average hourly earning, while covariates include: 1) dummy 
	variables---nonwhite, female, married, numdep, smsa, northcen, south, 
	west, 
	construc, ndurman, trcommpu, trade, services, profserv, profocc, clerocc, 
	and servocc; 2) non-dummy variables---educ, exper, and tenure; 3) interaction 
	variables---nonwhite $\times$ educ, nonwhite $\times$ exper, nonwhite 
	$\times$ tenure, female $\times$ educ, female $\times$ exper, female 
	$\times$ tenure, married $\times$ educ, married $\times$ exper, and married 
	$\times$ tenure.
	
	We consider the following two cases: 1) we rank the covariates according to 
	their linear correlations with the response variable, and then 
	consider the strictly nested model averaging method (intercept term is considered and ranked first); 2) 100 models are 
	selected by using function "regsubsets" in "leaps" package of R language, 
	where the parameters "nvmax" and "nbest" are taken to be 20 and 5 respectively, and other 
	parameters use default values.
	
\begin{figure}[ht]
		\centering
		\includegraphics[scale=0.9]{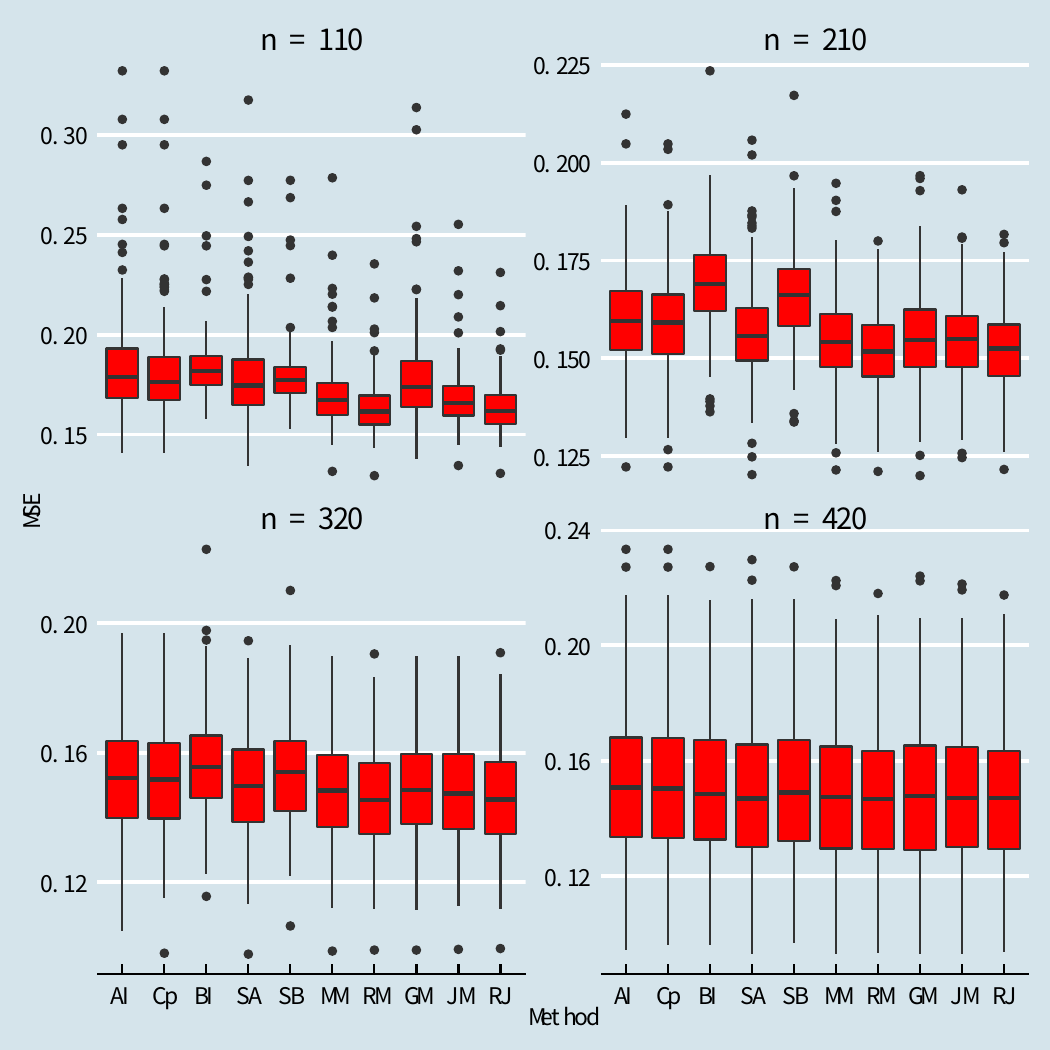}
		\caption{The box plot of MSE in Case 1 }
		\label{pic:13}
	\end{figure}

\begin{figure}[ht]
		\centering
		\includegraphics[scale=0.9]{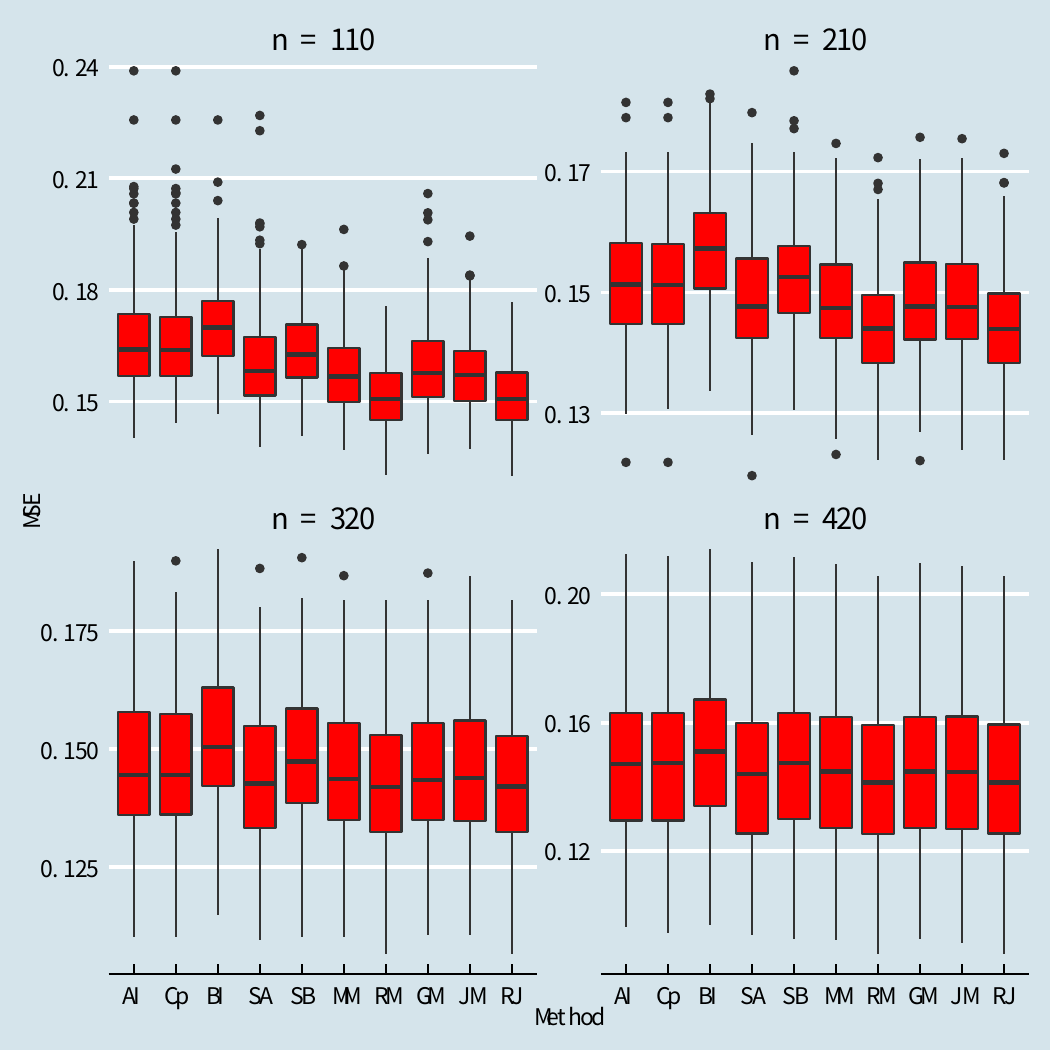}
		\caption{The box plot of MSE in Case 2 }
		\label{pic:14}
	\end{figure}

	We randomly divide the data into two parts: a training sample $S$ of $n$ 
	observations for estimating the models and a test sample $S_t$ of $n_t = 
	529 - n$ observations for validating the results. We consider $n = 110, 
	210, 320, 420$, and 
	$$ MSE = n_t^{-1} \sum_{ z_i \in  S_t }    [y_i - x_i^{'} \hat{\theta}(\hat{w})] ^2 $$
	is calculated as a measure of consistency. We replicate the process for 200 times. The box plots of MSEs in 200 simulations are shown 
	in Figures $\ref{pic:13}-\ref{pic:14}$. From these figures, we see that the performance of RM and RJ is good and stable. We also compute the 
	mean and median of MSEs, as well as the best performance rate (BPR), which 
	is the frequency of achieving the lowest risk across the replications. The 
	results are shown in Tables $\ref{tab:1}-\ref{tab:2}$. From these tables, we can draw the following conclusions:
	1) RM and RJ always are superior to other methods in terms of mean and median of MSEs, and BPR; 2) The performance of RM and RJ is basically the same in terms of mean and median of MSEs; 3) For BPR, on the whole, RM outperforms RJ.
	
	\begin{table}[htbp]
		\centering
		\caption{The mean, median and BPR of MSE in Case 1}
		\begin{tabular}{cccccccccccc}
			\toprule
			n     &       & AI    & Cp    & BI    & SA    & SB    & MM    & RM    & GM    & JM    & RJ \\
			\midrule
			\multirow{3}[2]{*}{110} & Mean  & 0.185  & 0.182  & 0.184  & 0.180  & 0.179  & 0.169  & 0.164  & 0.178  & 0.168  & 0.164  \\
			& Median & 0.179  & 0.176  & 0.182  & 0.174  & 0.177  & 0.167  & 0.162  & 0.174  & 0.166  & 0.162  \\
			& BPR   & 0.011  & 0.000  & 0.006  & 0.061  & 0.039  & 0.028  & 0.376  & 0.044  & 0.088  & 0.348  \\
			\midrule
			\multirow{3}[2]{*}{210} & Mean  & 0.160  & 0.159  & 0.169  & 0.157  & 0.166  & 0.155  & 0.152  & 0.155  & 0.155  & 0.152  \\
			& Median & 0.160  & 0.159  & 0.169  & 0.156  & 0.166  & 0.154  & 0.152  & 0.155  & 0.155  & 0.153  \\
			& BPR   & 0.030  & 0.000  & 0.010  & 0.075  & 0.000  & 0.020  & 0.460  & 0.080  & 0.035  & 0.290  \\
			\midrule
			\multirow{3}[2]{*}{320} & Mean  & 0.152  & 0.152  & 0.156  & 0.150  & 0.154  & 0.148  & 0.146  & 0.148  & 0.148  & 0.146  \\
			& Median & 0.152  & 0.152  & 0.156  & 0.150  & 0.154  & 0.148  & 0.145  & 0.148  & 0.147  & 0.146  \\
			& BPR   & 0.010  & 0.000  & 0.050  & 0.160  & 0.035  & 0.045  & 0.315  & 0.070  & 0.020  & 0.295  \\
			\midrule
			\multirow{3}[2]{*}{420} & Mean  & 0.152  & 0.152  & 0.150  & 0.150  & 0.151  & 0.148  & 0.147  & 0.149  & 0.148  & 0.147  \\
			& Median & 0.151  & 0.150  & 0.148  & 0.147  & 0.149  & 0.147  & 0.147  & 0.148  & 0.147  & 0.147  \\
			& BPR   & 0.025  & 0.005  & 0.105  & 0.175  & 0.060  & 0.030  & 0.215  & 0.055  & 0.030  & 0.300  \\
			\bottomrule
		\end{tabular}%
		\label{tab:1}%
	\end{table}%
	
	\begin{table}[htbp]
		\centering
		\caption{The mean, median and BPR of MSE in Case 2}
		\begin{tabular}{cccccccccccc}
			\toprule
			n     &       & AI    & Cp    & BI    & SA    & SB    & MM    & RM    & GM    & JM    & RJ \\
			\midrule
			\multirow{3}[2]{*}{110} & Mean  & 0.167  & 0.167  & 0.171  & 0.161  & 0.163  & 0.158  & 0.152  & 0.160  & 0.158  & 0.152  \\
			& Median & 0.164  & 0.164  & 0.170  & 0.158  & 0.163  & 0.157  & 0.151  & 0.158  & 0.157  & 0.151  \\
			& BPR   & 0.011  & 0.000  & 0.005  & 0.081  & 0.000  & 0.016  & 0.400  & 0.011  & 0.011  & 0.465  \\
			\midrule
			\multirow{3}[2]{*}{210} & Mean  & 0.152  & 0.151  & 0.157  & 0.149  & 0.153  & 0.148  & 0.145  & 0.148  & 0.148  & 0.145  \\
			& Median & 0.151  & 0.151  & 0.157  & 0.148  & 0.153  & 0.147  & 0.144  & 0.148  & 0.148  & 0.144  \\
			& BPR   & 0.000  & 0.005  & 0.000  & 0.170  & 0.010  & 0.005  & 0.460  & 0.005  & 0.015  & 0.330  \\
			\midrule
			\multirow{3}[2]{*}{320} & Mean  & 0.147  & 0.146  & 0.152  & 0.144  & 0.148  & 0.145  & 0.142  & 0.145  & 0.145  & 0.142  \\
			& Median & 0.145  & 0.145  & 0.150  & 0.143  & 0.147  & 0.144  & 0.142  & 0.143  & 0.144  & 0.142  \\
			& BPR   & 0.010  & 0.000  & 0.000  & 0.280  & 0.040  & 0.000  & 0.340  & 0.000  & 0.000  & 0.330  \\
			\midrule
			\multirow{3}[2]{*}{420} & Mean  & 0.148  & 0.148  & 0.152  & 0.144  & 0.148  & 0.146  & 0.143  & 0.146  & 0.146  & 0.143  \\
			& Median & 0.147  & 0.147  & 0.151  & 0.144  & 0.147  & 0.145  & 0.141  & 0.145  & 0.145  & 0.141  \\
			& BPR   & 0.005  & 0.000  & 0.000  & 0.370  & 0.055  & 0.000  & 0.265  & 0.000  & 0.010  & 0.295  \\
			\bottomrule
		\end{tabular}%
		\label{tab:2}%
	\end{table}%

\section{Concluding Remarks}
\label{sec6}
	In this paper, we study the relationship between AERM, stability, 
	generalization and consistency in model averaging. The results indicate 
	that stability is an important property of model averaging, which can 
	ensure that model averaging has good generalization performance and is 
	consistent under reasonable conditions. When the model weights are not 
	restricted, similar to ridge regression in \cite{RLSEl1970}, 
	extreme weights tend to appear under the influence of correlation between candidate
	models in MMA and JMA, resulting in poor performance of the corresponding model averaging 
	estimator. So we propose a $L_2$-penalty model averaging method. We prove 
	that it has stability and consistency. In order to reduce the impact of 
	tuning parameter selection, we use 10-fold cross-validation to 
	select a candidate set of tuning parameters and perform a weighted average 
	of the estimators of model weights based on estimation errors. The 
	numerical simulation and real data analysis show the superiority of the 
	proposed method.  
	
	Many issues deserve to be further investigated. We only apply the methods of Section $\ref{sec2}$ to the generalization of MMA and JMA in linear regression. It is worth investigating whether it is possible to extend the proposed method to more complex scenarios, such as generalized linear model, quantile regression, and dependent data. Further, we also expect to be able to propose a model averaging framework with stability and consistency that can be applied to multiple scenarios. In addition, in RMMA and RJMA, we can see that the estimators of weight vector are explicitly expressed. So, how to study their asymptotic behavior based on these explicit expressions is a meaningful but challenging topic.

\section{Acknowledgements}
\label{sec7}

	This work was partially supported by the National Natural Science Foundation of China (Grant nos. 11971323 and 12031016).

\appendix

\section{Lemmas and Proofs}
\label{sec9}

	Let $\hat{\theta}^{-n}(w)$ and $\hat{\theta}^{-(n-1,n)}(w)$ be the
	model averaging estimators corresponding to $\hat{\theta}(w)$, which are based on $S^{-n}$ and $S^{-(n-1,n)}$ 
	respectively, where $S^{-(n-1,n)}$ represents the sample set after removing 
	the $(n-1)$-th and $n$-th observations from $S$. 

	\begin{lemma}
		Under Assumptions \ref{ass:1} and \ref{ass:2}, there exists a constant $B_1>0$ such that $\mathop{max}\limits_{1\leq m \leq M} \|\hat{\theta}_m \|_2^2 \leq  B_1  K , a.s.$.
		\label{lem:1}
	\end{lemma}
	\begin{Proof}
		
		It follows from Assumptions \ref{ass:1} and \ref{ass:2} that 
		\begin{align*}
			& \max_{1\leq m \leq M} \|\hat{\theta}_m \|_2^2 \\
			&= \max_{1\leq m \leq M}  \|  (X_m^{'}X_m)^{-1} X_m^{'} y \|_2^2 \\
			&= \max_{1\leq m \leq M} y^{'} X_m(X_m^{'}X_m)^{-1} (X_m^{'}X_m)^{-1} X_m^{'} y \\
			&\leq  C_1^{-2} C_2 K n^{-1}  y^{'}y \\
			&\leq  C_1^{-2} C_2 C_3 K, a.s..
		\end{align*}
		We take $B_1 =C_1^{-2} C_2 C_3 $ to complete the proof.
		
	\end{Proof}
	\begin{lemma}
		Under Assumptions \ref{ass:1} and \ref{ass:2}, we have
		$$E_{S  } \Big ( \max_{1 \leq m \leq M} \|\hat{\theta}_m-\hat{\theta}^{-n}_m\|_2^2 \Big) =  O(n^{-2} 
		K^3) ,$$
		and
		$$E_{S  }  \Big (\max_{1 \leq m \leq M} \|\hat{\theta}_m^{-n}-\hat{\theta}^{-(n-1,n)}_m\|_2^2 \Big) =  O( n^{-2} 
		K^3 ).$$
		\label{lem:2}
	\end{lemma}
	\begin{Proof}
		
		By \cite{LOO1982}, we see that for any $m \in \{1,...,M\}$,
		$$\hat{\theta}_m =\hat{\theta}^{-n}_m + (X_m^{'} X_m)^{-1} \pi_m^{'} x_n (y_n - 
		x_n^{'} \pi_m \hat{\theta}^{-n}_m )  . $$
		It follows from Assumption \ref{ass:1} that
		\begin{align*}
			&E_{S   } \Big [\max_{1 \leq m \leq M}  \| \hat{\theta}_m 
			-\hat{\theta}^{-n}_m\|_2^2  \Big ]\\
			&=E_{S   } \Big [\max_{1 \leq m \leq M} x_n^{'}\pi_m (X_m^{'} X_m)^{-1} (X_m^{'} X_m)^{-1} \pi_m^{'} x_n (y_n - x_n^{'} \pi_m \hat{\theta}^{-n}_m )^2  \Big ]\\
			&\leq E_{S   } \Big [ \max_{1 \leq m \leq M} \lambda_{max}^2[(X_m^{'} X_m)^{-1}] \|\pi_m^{'} x_n (y_n - x_n^{'} \pi_m \hat{\theta}^{-n}_m )\|_2^2 \Big] \\
			&\leq  E_{S   }\Big[ C_1^{-2} n^{-2} \max_{1 \leq m \leq M} \|\pi_m^{'} x_n (y_n - x_n^{'} \pi_m
			\hat{\theta}^{-n}_m )\|_2^2\Big].  & \tag{1} \label{equ:1} 
		\end{align*}
		From Assumption \ref{ass:2}, we obtain 
		\begin{align*}
			& E_{S   } \Big[\max_{1 \leq m \leq M} \|\pi_m^{'} x_n (y_n - x_n^{'} \pi_m \hat{\theta}^{-n}_m )\|_2^2\Big] \\
			&\leq E_{S   } \Big[ \max_{1 \leq m \leq M}  \sum_{k=1}^{K}  x_{(k)n}^2 (y_n - x_n^{'} \pi_m \hat{\theta}^{-n}_m )^2\Big] \\
			&\leq C_4^2 K E_{S   } \Big[\max_{1 \leq m \leq M} (y_n - x_n^{'} \pi_m \hat{\theta}^{-n}_m )^2] \\
			&\leq C_4^2 K E_{S   } [ \max_{1 \leq m \leq M}(2y_n^2 + 2(x_n^{'} 
			\pi_m	 \hat{\theta}^{-n}_m )^2 )\Big] .   & \tag{2} \label{equ:2} 
		\end{align*}
		Further,  from Assumption \ref{ass:2} and Lemma $\ref{lem:1}$, we have
		\begin{align*}
			&E_{S   } \Big[\max_{1 \leq m \leq M} (x_n^{'} \pi_m \hat{\theta}^{-n}_m )^2 \Big]  \\
			&\leq  E_{S   } \Big (\max_{1 \leq m \leq M} \| x_n \|_2^2  \| \hat{\theta}^{-n}_m \|_2^2 \Big) \\
			&\leq C_4^2 K E_{S   }\Big  (\max_{1 \leq m \leq M} \| \hat{\theta}^{-n}_m \|_2^2 \Big) \\
			&\leq B_1 C_4^2 K^2   .   & \tag{3} \label{equ:3} 
		\end{align*}
		Combining $(\ref{equ:1})-(\ref{equ:3})$, it is seen that
		$$E_{S  } \Big  (\max_{1 \leq m \leq M} \|\hat{\theta}_m-\hat{\theta}^{-n}_m\|_2^2 \Big) =  O( n^{-2} 
		K^3) .$$
		In a similar way, it can be shown that
		$$E_{S  } \Big  (\max_{1 \leq m \leq M} \|\hat{\theta}_m^{-n}-\hat{\theta}^{-(n-1,n)}_m\|_2^2 \Big) =  O( n^{-2} 
		K^3 ).$$
		
	\end{Proof}
	\begin{lemma}
		Under Assumptions \ref{ass:1} and \ref{ass:2}, we have
		$$ E_{S   ,z^* } (\hat{\gamma}^{'} \hat{\gamma})  = n^{-1}E_{S   } (\hat{\Omega}^{'} \hat{\Omega} ) + O( n^{-1}K^3 ) ,$$
		and
		$$ n^{-1}E_{S   } (\bar{\Omega}^{'} \bar{\Omega})   = n^{-1}E_{S   } (\hat{\Omega}^{'} \hat{\Omega} )   + O[ n (C_1n -C_4^2 K)^{-2}  K] .$$
		\label{lem:3}
	\end{lemma}
	\begin{Proof}
		
		Note that
		$$ \hat{\gamma}^{'} \hat{\gamma} =  ( x^{*'} \pi_m \hat{\theta}_m   \hat{\theta}_t^{'} \pi_t^{'} x^{*} )_{M \times M},$$
		$$ \bar{\Omega}^{'} \bar{\Omega} =\big\{ [y-D_m(y-P_my) ]^{'}[y-D_t(y-P_ty)] \big\}_{M \times M},  $$
		and
		$$ \hat{\Omega}^{'} \hat{\Omega} =( y^{'}P_m^{'}P_ty  )_{M \times M}  =  \Big ( \sum_{i=1}^n x_i^{'} \pi_m \hat{\theta}_m \hat{\theta}_t^{'} \pi_t^{'}   x_i  \Big)_{M \times M}.$$
		
		It follows from Assumption \ref{ass:2}, Lemma $\ref{lem:1}$ and Lemma $\ref{lem:2}$ that
		\begin{align*}
			&|E_{S  }   ( x_n^{'} \pi_m \hat{\theta}_m \hat{\theta}_t^{'} \pi_t^{'}   x_n  -x_n^{'} \pi_m \hat{\theta}_m^{-n}   \hat{\theta}_t^{-n'} \pi_t^{'} x_n )| \\
			&\leq |E_{S  }    [x_n^{'} \pi_m (\hat{\theta}_m - \hat{\theta}_m^{-n}) \hat{\theta}_t^{'} \pi_t^{'}   x_n] |+|E_{S  }   [x_n^{'} \pi_m \hat{\theta}_m^{-n} (\hat{\theta}_t-   \hat{\theta}_t^{-n})^{'}  \pi_t^{'} x_n ]| \\
			&\leq \sqrt{E_{S  }( \| \hat{\theta}_m - \hat{\theta}_m^{-n}\|_2^2 
				)  }\sqrt{E_{S  }(\|x_n^{'} \pi_t \hat{\theta}_t \pi_m^{'}x_n 
				\|_2^2) } +   \sqrt{E_{S  }(\|x_n^{'} \pi_m \hat{\theta}_m^{-n} 
				\pi_t^{'}x_n \|_2^2)}\sqrt{E_{S  }(\|\hat{\theta}_t-   
				\hat{\theta}_t^{-n} \|_2^2)} \\
			&\leq \sqrt{E_{S  }( \| \hat{\theta}_m - \hat{\theta}_m^{-n}\|_2^2 
				)  }\sqrt{E_{S  }(\|x_n \|_2^2 \|\hat{\theta}_t \|_2^2 \|x_n 
				\|_2^2) } + \sqrt{E_{S  }(\|x_n \|_2^2 \|\hat{\theta}_m^{-n} \|_2^2 
				\|x_n \|_2^2)}\sqrt{E_{S  }(\|\hat{\theta}_t-   \hat{\theta}_t^{-n} 
				\|_2^2)} \\
			&=O( n^{-1} K^3).
		\end{align*}
		In a similar way, we obtain
		$$|E_{S   ,  z^*  }  (x^{*'} \pi_m \hat{\theta}_m^{-n}   \hat{\theta}_t^{-n'} \pi_t^{'} x^{*} - x^{*'} \pi_m \hat{\theta}_m   \hat{\theta}_t^{'} \pi_t^{'} x^{*})|=O( n^{-1} K^3).$$
		Further, it is seen that
		\begin{align*}
			& E_{S   , z^*  }  (\frac{1}{n} \sum_{i=1}^n x_i^{'} \pi_m \hat{\theta}_m \hat{\theta}_t^{'} \pi_t^{'}   x_i - x^{*'} \pi_m \hat{\theta}_m   \hat{\theta}_t^{'} \pi_t^{'} x^{*})  \\
			&= \frac{1}{n} \sum_{i=1}^n E_{S   , z^*  }  ( x_i^{'} \pi_m \hat{\theta}_m \hat{\theta}_t^{'} \pi_t^{'}   x_i - x^{*'} \pi_m \hat{\theta}_m   \hat{\theta}_t^{'} \pi_t^{'} x^{*})  \\
			&=E_{S   , z^*  }  ( x_n^{'} \pi_m \hat{\theta}_m \hat{\theta}_t^{'} \pi_t^{'}   x_n - x^{*'} \pi_m \hat{\theta}_m   \hat{\theta}_t^{'} \pi_t^{'} x^{*})  \\
			&= E_{S   ,  z^*  }  ( x_n^{'} \pi_m \hat{\theta}_m 
			\hat{\theta}_t^{'} \pi_t^{'}   x_n  -x^{*'} \pi_m 
			\hat{\theta}_m^{-n}   \hat{\theta}_t^{-n'} \pi_t^{'} x^{*} )  
			+E_{S   ,z^*  }  (x^{*'} \pi_m \hat{\theta}_m^{-n}   
			\hat{\theta}_t^{-n'} \pi_t^{'} x^{*} - x^{*'} \pi_m 
			\hat{\theta}_m   \hat{\theta}_t^{'} \pi_t^{'} x^{*})  \\
			&=E_{S   }  ( x_n^{'} \pi_m \hat{\theta}_m \hat{\theta}_t^{'} 
			\pi_t^{'}   x_n  -x_n^{'} \pi_m \hat{\theta}_m^{-n}   
			\hat{\theta}_t^{-n'} \pi_t^{'} x_n )  +E_{S   ,  z^*  }  (x^{*'} 
			\pi_m \hat{\theta}_m^{-n}   \hat{\theta}_t^{-n'} \pi_t^{'} x^{*} - 
			x^{*'} \pi_m \hat{\theta}_m   \hat{\theta}_t^{'} \pi_t^{'} x^{*}) . 
		\end{align*}
		So we have
		$$ E_{S   ,z^*} (\hat{\gamma}^{'} \hat{\gamma})  = n^{-1} E_{S   } [\hat{\Omega}^{'} \hat{\Omega} ] + O( n^{-1}K^3 ) .$$
		
		On the other hand, it follows from Assumptions  $\ref{ass:1}$ and $\ref{ass:2}$ that
		$$ \max_{1 \leq i \leq n} \max_{1 \leq m \leq M} h_{ii}^m =  x_i^{'} \pi_m (X_m^{'} X_m)^{-1} \pi_m^{'} x_i \leq \frac{C_4^2 K}{nC_1} , a.s..$$
		Hence, we have
		\begin{align*}
			& | y^{'} P_m^{'} P_t y -  [y-D_m(y-P_my) ]^{'}[y-D_t(y-P_ty)]     |  \\
			&=|y^{'} P_m^{'} P_t y -  [(I_n-D_m)y+D_mP_my] ^{'}[(I_n-D_t)y+D_tP_ty ] |     \\ 
			&\leq |y^{'} P_m^{'} (D_m D_t - I_n  )  P_t y| +  y^{'}(I_n-D_m)(I_n-D_t)y  +|y^{'}(I_n-D_m)D_tP_ty | + |y^{'}P_m^{'}D_m(I_n-D_t)y| \\
			&\leq \max_{1 \leq m \leq M}  \max_{1 \leq i \leq n} [(1 -h_{ii}^m )^{-2}- 1]   y^{'} P_m^{'}  P_m y + \max_{1 \leq m \leq M} \max_{1 \leq i \leq n}  [1 -  (1 -h_{ii}^m )^{-1} ]^2 y^{'}y \\
			&\ \ \ +2 \max_{1 \leq m \leq M} \max_{1 \leq i \leq n} \sqrt{ (1-h_{ii}^{m})^{-2} [1-(1 -h_{ii}^m )^{-1}  ]^2 y^{'}y y^{'} P_m^{'}   P_m y  } \\
			&\leq \max_{1 \leq m \leq M} \max_{1 \leq i \leq n}\{ [(1 -h_{ii}^m )^{-2}-1 ] +   [ (1 -h_{ii}^m )^{-1}-1] ^2  + 2 (1-h_{ii}^{m})^{-1}   [(1 -h_{ii}^m )^{-1}-1]   \} y^{'}y \\
			&\leq C_3 n \{[ (1 - \frac{C_4^2 K}{nC_1} )^{-2}-1 ] +    [(1 - \frac{C_4^2 K}{nC_1} )^{-1}-1] ^2 + 2  (1 - \frac{C_4^2 K}{nC_1} )^{-1}  [ (1 - \frac{C_4^2 K}{nC_1} )^{-1}-1]  \}   \\
			&= \frac{4C_1 C_3 C_4^2 n^2 K  }{(C_1n -C_4^2 K)^2 } , a.s..
		\end{align*}
		Thus, from Assumption 2, we see that
		$$ n^{-1}E_{S } (\bar{\Omega}^{'} \bar{\Omega})  = n^{-1} E_{S   } (\hat{\Omega}^{'} \hat{\Omega})  + O[ n (C_1n -C_4^2 K)^{-2}  K] .$$
		
	\end{Proof}
	\begin{lemma}
		Under Assumptions \ref{ass:1} $-$ \ref{ass:3}, there is a constant $B_2>0$ such that
		$$ \|\hat{w} \|_2^2 \leq B_2 M (1+ n^{-2} K^2 ),$$
		$$ \|\bar{w} \|_2^2 \leq B_2 M ,$$
		$$ \|\tilde{w} \|_2^2 \leq B_2 M (1+ n^{-2} K^2 ),$$
		and
		$$ \|\hat{w}^* \|_2^2 \leq B_2 M^2, $$
		a.s..
		\label{lem:4}
	\end{lemma}
	\begin{Proof}
		
		It follows from Assumptions  $\ref{ass:2}$ and $\ref{ass:3}$ that
		\begin{align*}
			&\| \hat{w} \|_2^2 \\
			&=\| ( \hat{\Omega}^{'}\hat{\Omega} + \lambda_n I_n 
			)^{-1}\hat{\Omega}^{'} y   - \sigma^2  ( 
			\hat{\Omega}^{'}\hat{\Omega} + \lambda_n I_n)^{-1}  \kappa \|_2^2 \\
			&\leq 2 y^{'} \hat{\Omega} ( \hat{\Omega}^{'}\hat{\Omega} + 
			\lambda_n I_n )^{-1}( \hat{\Omega}^{'}\hat{\Omega} + \lambda_n I_n 
			)^{-1} \hat{\Omega}^{'} y  +2\sigma^4  \kappa^{'} ( 
			\hat{\Omega}^{'}\hat{\Omega} + \lambda_n I_n )^{-1}( 
			\hat{\Omega}^{'}\hat{\Omega} + \lambda_n I_n )^{-1}  \kappa   \\
			&\leq 2C_5^{-2} C_6 n^{-1} M y^{'}y    + 2\sigma^4 C_5^{-2}  
			n^{-2}  
			\kappa^{'} \kappa  \\
			&\leq 2C_5^{-2}  C_3 C_6 M    + 2\sigma^4 C_5^{-2}  n^{-2}  K^2 M , 
			a.s.,
		\end{align*}
		and 
		\begin{align*}
			&\| \bar{w} \|_2^2 \\
			&=\|( \bar{\Omega}^{'}\bar{\Omega}+ \lambda_n I_n 
			)^{-1}\bar{\Omega}^{'} y   \|_2^2 \\
			&\leq  y^{'} \bar{\Omega} ( \bar{\Omega}^{'}\bar{\Omega}+ \lambda_n 
			I_n )^{-1}( \bar{\Omega}^{'}\bar{\Omega}+ \lambda_n I_n )^{-1} 
			\bar{\Omega}^{'} y  \\
			&\leq C_5^{-2} C_{6} n^{-1} M y^{'}y     \\
			&\leq C_5^{-2}  C_3 C_{6}  M  , a.s.. 
		\end{align*}
		In a similar way, we obtain
		$$ \|\tilde{w}\|_2^2 \leq 2C_5^{-2}  C_3 C_6 M    + 
		2\sigma^4 C_5^{-2}  n^{-2}  K^2 M, a.s.. $$
		On the other hand, it follows from Assumptions  $\ref{ass:2}-\ref{ass:3}$ and Lemma $\ref{lem:1}$ that  
		\begin{align*}
			&\| \hat{w}^{*} \|_2^2 \\
			&=\| [E_{z^*  } (\hat{\gamma}^{'} \hat{\gamma})] ^{-1}E_{z^*  } 
			(\hat{\gamma}^{'} y^*) \|_2^2 \\
			&\leq E_{z^*  } (\hat{\gamma} y^*)   [E_{z^*  } (\hat{\gamma}^{'} 
			\hat{\gamma})] ^{-1} [E_{z^*  } (\hat{\gamma}^{'} 
			\hat{\gamma})] ^{-1} E_{z^*  } (\hat{\gamma}^{'} y^*)  \\
			&\leq E_{z^*  } (\hat{\gamma} y^*) E_{z^*  } (\hat{\gamma}^{'} 
			y^*)   \lambda_{min}^{-2}[E_{z^*  } (\hat{\gamma}^{'} 
			\hat{\gamma})]    \\
			&\leq C_5^{-2}  E_{z^*  } (\hat{\gamma} y^*) E_{z^*  } 
			(\hat{\gamma}^{'} y^*) \\
			&\leq C_5^{-2}  E_{z^*  } (\hat{\gamma}\hat{\gamma}^{'}) E_{z^*  } 
			(y^{*2})   \\
			&\leq C_5^{-2}  tr[ E_{z^*  } (\hat{\gamma}^{'}\hat{\gamma})]  
			E_{z^*  } (y^{*2})   \\
			&\leq C_5^{-2}  C_3 C_6 M^2  , a.s..
		\end{align*}
		We take $B_2 = max\{ 2C_5^{-2}  C_3 C_6     ,2\sigma^4 C_5^{-2} \}$ to 
		complete the proof.
		
	\end{Proof}
	\begin{lemma}
		Under Assumptions \ref{ass:1} $-$ \ref{ass:4}, we have
		$$ E_{S  } [C( \hat{w}^{-n},S ) - C(\hat{w},S) ] = O[ K^3   M^2( 
		1+n^{-2}K^2 ) ],$$
		$$ E_{S  } [J( \bar{w}^{-n},S ) - J(\bar{w},S) ] = O( K^3   M^2),$$
		and
		$$ E_{S  } [F( \hat{w}^{*,-n},S ) - F(\hat{w}^*,S) ] = O(n^{-1} K^3 M^3 
		).$$
		\label{lem:5}
	\end{lemma}
	\begin{Proof}
		
		Note that
		\begin{align*}
			&C(\hat{w}^{-n},S)- C(\hat{w},S) \\
			&=C(\hat{w}^{-n},S)-C(\hat{w}^{-n}, S^{-n} )+C(\hat{w}^{-n},S^{-n})-C(\hat{w},S) \\
			&\leq C(\hat{w}^{-n},S)-C(\hat{w}^{-n}, S^{-n} )+C(\hat{w},S^{-n})-C(\hat{w},S),  
		\end{align*}
		and 
		\begin{align*}
			& E_{S   } [C(w,S)-C(w,S^{-n})- (\lambda_n - \lambda_{n-1} )w^{'}w]   \\
			&= \sum_{i=1}^{n-1} E_{S   } \Big\{ [y_i-x_i^{'} \hat{\theta}(w) ]^2 - 
			[y_i-x_i^{'} \hat{\theta}^{-n}(w)] ^2 \Big\} + E_{S   } \Big\{[y_n-x_n^{'} \hat{\theta}(w) ]^2\Big\} \\
			&= (n-1) E_{S   } \Big\{[y_{n-1}-x_{n-1}^{'} \hat{\theta}(w) ]^2 - 
			[y_{n-1}-x_{n-1}^{'} \hat{\theta}^{-n}(w) ]^2\Big\}  + E_{S   } \Big\{[y_n-x_n^{'} \hat{\theta}(w) ]^2\Big\},
		\end{align*}
		where 
		$$\hat{w}^{-n} = argmin_{w \in R^M} C(w,S^{-n}),$$
		and
		$$ C(w,S^{-n}) =  \sum_{i=1}^{n-1} [ y_i -  x_i^{'}  \hat{\theta}^{-n}(w)]^2 + 2 \sigma^2 w^{'} \kappa + \lambda_{n-1} w^{'}w.$$
		It follows from Lemmas  $\ref{lem:1}$, $\ref{lem:2}$ and $\ref{lem:4}$ that
		\begin{align*}
			& \|\hat{\theta}(\hat{w})\|_2^2  \\
			&= \sum_{m=1}^M \sum_{t=1}^M \hat{w}_m \hat{w}_t \hat{\theta}_m^{'} \pi_m^{'} \pi_t \hat{\theta}_t\\
			&\leq \sum_{m=1}^M \sum_{t=1}^M |\hat{w}_m \hat{w}_t| 
			|\hat{\theta}_m^{'} \pi_m^{'} \pi_t \hat{\theta}_t|\\
			&\leq \max_{1 \leq m \leq M} \max_{1 \leq t \leq M} 
			|\hat{\theta}_m^{'} \pi_m^{'} \pi_t \hat{\theta}_t| \sum_{m=1}^M 
			\sum_{t=1}^M |\hat{w}_m \hat{w}_t| \\
			&\leq \max_{1 \leq m \leq M} \max_{1 \leq t \leq M} 
			\|\hat{\theta}_m\|_2 \|\hat{\theta}_t\|_2  \Big( \sum_{m=1}^M 
			|\hat{w}_m | \Big)^2\\
			&= M \max_{1 \leq m \leq M}  \|\hat{\theta}_m\|_2^2  \sum_{m=1}^M 
			\hat{w}_m^2 \\ 
			&\leq B_1 B_2 K M^2( 1+n^{-2}K^2), a.s.,
		\end{align*}
		and
		\begin{align*}
			& E_{S   } [\|\hat{\theta}(\hat{w}) -\hat{\theta}^{-n}(\hat{w})\|_2^2] \\
			&= E_{S   } \Big[\sum_{m=1}^M \sum_{t=1}^M \hat{w}_m \hat{w}_t (\hat{\theta}_m -\hat{\theta}_m^{-n} )^{'}  \pi_m^{'} \pi_t (\hat{\theta}_t -\hat{\theta}_t^{-n} )  \Big] \\
			&= E_{S   }\Big( \sum_{m=1}^M \sum_{t=1}^M |\hat{w}_m \hat{w}_t| |(\hat{\theta}_m -\hat{\theta}_m^{-n} )^{'}  \pi_m^{'} \pi_t (\hat{\theta}_t -\hat{\theta}_t^{-n} )| \Big)  \\
			&\leq E_{S   } \Big(\sum_{m=1}^M \sum_{t=1}^M |\hat{w}_m \hat{w}_t| 
			\|\hat{\theta}_m -\hat{\theta}_m^{-n} \|_2 \|\hat{\theta}_t 
			-\hat{\theta}_t^{-n} \|_2\Big)  \\
			&\leq M E_{S   } \Big(\max_{1 \leq m \leq M} \max_{1 \leq t \leq M} 
			\|\hat{\theta}_m -\hat{\theta}_m^{-n} \|_2  \|\hat{\theta}_t 
			-\hat{\theta}_t^{-n} \|_2 \sum_{m=1}^M \hat{w}_m^2 \Big)  \\
			&\leq B_2 M^2( 1+n^{-2}K^2 ) E_{S   }  \Big(\max_{1 \leq m \leq M} 
			\max_{1 \leq t \leq M} \|\hat{\theta}_m -\hat{\theta}_m^{-n} \|_2  
			\|\hat{\theta}_t -\hat{\theta}_t^{-n} \|_2  \Big)\\
			&=B_2 M^2( 1+n^{-2}K^2 ) E_{S   }  \Big(\max_{1 \leq m \leq M}  
			\|\hat{\theta}_m -\hat{\theta}_m^{-n} \|_2^2 \Big) \\
			&\leq C_8 B_2 n^{-2} K^3 M^2( 1+n^{-2}K^2 ) .
		\end{align*}
		In a similar way, we obtain
		$$\| \hat{\theta}^{-n}(\hat{w})\|_2^2 \leq B_1 B_2 K M^2( 1+n^{-2}K^2), 
		a.s..$$
		Further, from Assumption $\ref{ass:2}$, we have
		\begin{align*}
			&  E_{S   } \Big\{ \|x_{n-1}[2y_{n-1} - x_{n-1}^{'} 
			\hat{\theta}(\hat{w}) - x_{n-1}^{'} \hat{\theta}^{-n}(\hat{w})] \|_2^2 \Big\} \\
			&= \sum_{k=1}^{K} E_{S   } \Big\{x_{{n-1}k}^2 [2y_{n-1} - 
			x_{n-1}^{'} \hat{\theta}(\hat{w}) - x_{n-1}^{'} \hat{\theta}^{-n}(\hat{w})]^2\Big\} 
			\\
			&\leq  C_4^2 K E_{S   } \Big\{ [2y_{n-1} - x_{n-1}^{'} 
			\hat{\theta}(\hat{w}) - x_{n-1}^{'} \hat{\theta}^{-n}(\hat{w})]^2\Big\} \\
			&\leq  C_4^2 K E_{S   }  \Big\{ 6y_{n-1}^2 + 3[ x_{n-1}^{'} 
			\hat{\theta}(\hat{w}) ]^2 +3[x_{n-1}^{'} \hat{\theta}^{-n}(\hat{w})]^2 \Big\}  \\
			&\leq  C_4^2 K E_{S   }  [6y_{n-1}^2 + 3\|x_{n-1}\|_2^2 
			\|\hat{\theta}(\hat{w})\|_2^2 +3 \|x_{n-1}\|_2^2 \| \hat{\theta}^{-n}(\hat{w}) \|_2^2]  \\
			&\leq 6 C_3 C_4^2 K + 6 C_4^4 B_1 B_2 K^3   M^2( 1+n^{-2}K^2) ,
		\end{align*}
		and
		\begin{align*}
			& E_{S   } \{ [y_n-x_n^{'} \hat{\theta}(\hat{w} ) ]^2\} \leq 2 C_3  
			+ 2 C_4^2 B_1 B_2 K^2   M^2( 1+n^{-2}K^2 ).
		\end{align*}
		These arguments indicate that
		\begin{align*}
			&  |E_{S   }\Big\{   [y_{n-1}-x_{n-1}^{'} \hat{\theta}(\hat{w})] ^2 - [y_{n-1}-x_{n-1}^{'} \hat{\theta}^{-n} (\hat{w})]^2 | \Big\} \\
			&= |E_{S   } \Big\{[2y_{n-1} - x_{n-1}^{'} \hat{\theta}(\hat{w}) - x_{n-1}^{'} \hat{\theta}^{(-n)}(\hat{w})][x_{n-1}^{'} \hat{\theta}(\hat{w}) - x_{n-1}^{'} \hat{\theta}^{-n}(\hat{w}) 
			] \Big\}| \\
			&\leq \sqrt{E_{S   }  [\|x_{n-1}(2y_{n-1} - x_{n-1}^{'} \hat{\theta}(\hat{w}) - x_{n-1}^{'} \hat{\theta}^{-n}(\hat{w}) \|_2^2] }  
			\sqrt{E_{S   } [\|\hat{\theta}(\hat{w}) 
				-\hat{\theta}^{-n}(\hat{w})\|_2^2] } \\
			&=O [ n^{-1} K^3   M^2( 1+n^{-2}K^2 ) ],
		\end{align*}
		and
		$$|E_{S   } [C(\hat{w},S^{-n})- C(\hat{w},S) - (\lambda_n - 
		\lambda_{n-1} )\hat{w}^{'}\hat{w}]| = O[ K^3   M^2( 1+n^{-2}K^2 ) ] .$$
		Similarly, we obtain
		$$|E_{S   } 
		[C(\hat{w}^{-n},S)-C(\hat{w}^{-n},S^{-n}) -(\lambda_n - \lambda_{n-1} 
		)\hat{w}^{-n'}\hat{w}^{-n} ]| = O [ K^3   M^2( 1+n^{-2}K^2) ] .$$
		Thus, it follows from Assumption $\ref{ass:4}$ and Lemma $\ref{lem:4}$ that
		$$E_{S   }[ C(\hat{w}^{-n},S) - C(\hat{w},S)]  = O[ K^3   M^2( 
		1+n^{-2}K^2 ) ] .$$
		
		On the other hand, we notice that
		\begin{align*}
			&J(\bar{w}^{-n},S)- J(\bar{w},S) \\
			&=J(\bar{w}^{-n},S)-J(\bar{w}^{-n},S^{-n})+J(\bar{w}^{-n},S^{-n})-J(\bar{w},S) \\
			&\leq J(\bar{w}^{-n},S)-J(\bar{w}^{-n},S^{-n})+J(\bar{w},S^{-n})-J(\bar{w},S),
		\end{align*}
		and
		\begin{align*}
			& E_{S   } [J(w,S)-J(w,S^{-n}) - (\lambda_n - \lambda_{n-1} )w^{'}w ] \\
			&= \sum_{i=1}^{n-1} E_{S   }\Big\{ [y_i-x_i^{'} \hat{\theta}^{-i}(w) ]^2 - 
			[y_i-x_i^{'} \hat{\theta}^{-(i,n)}(w) ]^2 \Big\} + E_{S   } \Big\{ [y_n-x_n^{'} \hat{\theta}^{-n}(w)] ^2\Big\} \\
			&= (n-1) E_{S   }\Big\{   [y_{n-1}-x_{n-1}^{'} \hat{\theta}^{-(n-1)}(w) ]^2 - 
			[y_{n-1}-x_{n-1}^{'} \hat{\theta}^{-(n-1,n)}(w)] ^2\Big \} \\
			&+ E_{S   } \Big\{  [y_n-x_n^{'} \hat{\theta}^{-n}(w) ]^2 \Big\},
		\end{align*}
		where 
		$$\bar{w}^{-n} = argmin_{w \in R^M} J(w,S^{-n}),$$
		and
		$$ J(w,S^{-n}) =  \sum_{i=1}^{n-1}  [y_i -  x_i^{'}  \hat{\theta}^{-(i,n)}(w)]^2  + \lambda_{n-1} w^{'}w.$$
		So, we obtain
		$$|E_{S   } [J(\bar{w},S^{-n})- J(\bar{w},S)- (\lambda_n - 
		\lambda_{n-1} )\bar{w}^{'}\bar{w}]| = O(  K^3   M^2) ,$$
		and
		$$|E_{S   } 
		[J(\bar{w}^{-n},S)-J(\bar{w}^{-n},S^{-n})- (\lambda_n - \lambda_{n-1} 
		)\bar{w}^{-n'}\bar{w}^{-n}]| = O(   K^3   M^2) .$$
		Further, we have
		$$E_{S   }[ J(\bar{w}^{-n},S) - J(\bar{w},S)]  = O(  K^3   M^2)  .$$
		
		Similarly, from
		\begin{align*}
			&E_{S  }[F(\hat{w}^{*,-n},S) -F(\hat{w}^{*},S)] \\
			&= E_{S  }[F(\hat{w}^{*,-n},S)  - F(\hat{w}^{*,-n},S^{-n})+ F(\hat{w}^{*,-n},S^{-n}) -F(\hat{w}^{*},S)] \\
			&\leq E_{S  } [F(\hat{w}^{*,-n},S)  - F(\hat{w}^{*,-n},S^{-n})+ F(\hat{w}^{*},S^{-n}) -F(\hat{w}^{*},S)]  ,
		\end{align*}
		and
		\begin{align*}
			& E_{S  } [ F(w,S^{-n}) -F(w,S)]  =E_{S  ,z^* }\Big\{  [y^* -  x^{*'} \hat{\theta}^{-n}(w) ]^2 -[y^* -  x^{*'}  \hat{\theta}(w) ]^2\Big \} ,
		\end{align*}
		where 
		$$\hat{w}^{*,-n} = argmin_{w \in R^M} F(w,S^{-n}),$$
		and
		$$ F(w,S^{-n}) =  E_{z^*   }\{ [y^* -  x^{*'}  \hat{\theta}^{-n}(w)]^2 \},$$
		we have
		$$E_{S   }[F(\hat{w}^{*,-n},S) -F(\hat{w}^{*},S)] =O(
		n^{-1} K^3 M^3) .$$

	\end{Proof}
	\begin{lemma}
		Under Assumptions \ref{ass:1} $-$ \ref{ass:4}, we have
		$$E_{S    }[\|\hat{w} - \hat{w}^{-n}\|_2^2] = O[ n^{-1} K^3   M^2( 
		1+n^{-2}K^2 ) ],$$
		$$E_{S    }[\|\bar{w} - \bar{w}^{-n}\|_2^2] = O( n^{-1} K^3   M^2) ,$$
		and
		$$E_{S    }[\|\hat{w}^* - \hat{w}^{*,-n}\|_2^2] = O( n^{-1} K^3   M^3 ) 
		.$$
		\label{lem:6}
	\end{lemma}
	\begin{Proof}
		
		It follows from the definition of $\hat{w}$ that $\frac{\partial C(\hat{w},S)}{\partial w } ={\bf{0}}_M$. So, we have
		\begin{align*}
			&C( \hat{w}^{-n},S ) - C( \hat{w},S )  \\
			&=  (\hat{w}^{-n} - \hat{w})^{'} \frac{\partial C(\hat{w},S)}{\partial w } + (\hat{w}^{-n} - \hat{w})^{'} (\hat{\Omega}^{'}\hat{\Omega} + \lambda_n I_n) (\hat{w}^{-n} - \hat{w}) \\
			&=(\hat{w}^{-n} - \hat{w})^{'} (\hat{\Omega}^{'}\hat{\Omega}+ \lambda_n I_n) (\hat{w}^{-n} - \hat{w})  \\
			&\geq  C_5 n \|\hat{w}^{-n} - \hat{w} \|_2^2, a.s..
		\end{align*}
		Further, it follows from Lemma \ref{lem:5} that
		$$ E_{S   } ( \|\hat{w}^{-n} - \hat{w} \|_2^2)  = O [n^{-1} K^3   M^2( 
		1+n^{-2}K^2 ) ]   .$$
		On the other hand, it follows from the definition of $\bar{w}$ that $\frac{\partial 
			J(\bar{w},S)}{\partial w } ={\bf{0}}_M$. So, we have
		\begin{align*}
			&J( \bar{w}^{-n},S ) - J( \bar{w},S )  \\
			&=  (\bar{w}^{-n} - \bar{w})^{'} \frac{\partial J(\bar{w},S)}{\partial w } + (\bar{w}^{-n} - \bar{w})^{'} (\bar{\Omega}^{'}\bar{\Omega} + \lambda_n I_n) (\bar{w}^{-n} - \bar{w}) \\
			&=(\bar{w}^{-n} - \bar{w})^{'} (\bar{\Omega}^{'}\bar{\Omega}+ \lambda_n I_n) (\bar{w}^{-n} - \bar{w})  \\
			&\geq C_5 n \|\bar{w}^{-n} - \bar{w} \|_2^2, a.s..
		\end{align*}
		Further, from Lemma \ref{lem:5}, we have
		$$ E_{S   } ( \|\bar{w}^{-n} - \bar{w} \|_2^2)  = O( n^{-1} K^3   M^2 
		)  .$$
		Similarly, from the definition of $\hat{w}^*$, Lemma \ref{lem:5} and
		\begin{align*}
			&F( \hat{w}^{*,-n},S ) - F( \hat{w}^*,S )  \\
			&=  (\hat{w}^{*,-n} - \hat{w}^*)^{'} \frac{\partial F(\hat{w}^*,S)}{\partial w } + (\hat{w}^{*,-n} - \hat{w}^*)^{'} E_{z^*  }( \hat{\gamma}^{'}\hat{\gamma} )  (\hat{w}^{*,-n} - \hat{w}^*) \\
			&=(\hat{w}^{*,-n} - \hat{w}^*)^{'} E_{z^*  }( \hat{\gamma}^{'}\hat{\gamma} ) (\hat{w}^{*,-n} - \hat{w}^*)  \\
			&\geq C_5  \|\hat{w}^{*,-n} - \hat{w}^* \|_2^2, a.s.,
		\end{align*}
		we obtain
		$$   E_{S   } (\|\hat{w}^{*,-n} - \hat{w}^* \|_2^2) = O(  n^{-1} K^3   
		M^3 ) .$$
		
	\end{Proof}
	
	\noindent {\bf{Proof of Theorem $\ref{thm:4-2-1}$:}}
	Let $(\hat{c}_1,...,\hat{c}_M)^{'} = \hat{P}^{'}\hat{w}^0,$ and  
	$(\hat{d}_1,...,\hat{d}_M)^{'} = \hat{P}^{'}\hat{w}^* $. Then, we have
	\begin{align*}
		&\hat{M}_1(\lambda_n) \\
		&= \|\hat{Z} \hat{w}^0 - \hat{Z} \hat{w}^* \|_2^2  +\|\hat{Z} \hat{w}^* -  \hat{w}^* \|_2^2  \\
		&=\sum_{m=1}^M \frac{ (\hat{c}_m-\hat{d}_m)^2 \hat{\zeta}_m^2 }{ (\lambda_n + \hat{\zeta}_m)^2 }  + \sum_{m=1}^M \hat{d}_m^2 (\frac{\hat{\zeta}_m}{\lambda_n + \hat{\zeta}_m} -1)^2   \\
		&= \sum_{m=1}^M \frac{ (\hat{c}_m-\hat{d}_m)^2 \hat{\zeta}_m^2 }{ (\lambda_n + \hat{\zeta}_m)^2 }  + \sum_{m=1}^M  \frac{\hat{d}_m^2 \lambda_n^2 }{(\lambda_n + \hat{\zeta}_m)^2 }   ,\\
	\end{align*}
	and
	\begin{align*}
		&\frac{d}{d \lambda_n} \hat{M}_1(\lambda_n) \\
		&= \sum_{m=1}^M \frac{ -2(\hat{c}_m-\hat{d}_m)^2 \hat{\zeta}_m^2  }{ (\lambda_n + \hat{\zeta}_m)^3 }  + \sum_{m=1}^M  \frac{2 \hat{d}_m^2 \lambda_n (\lambda_n + \hat{\zeta}_m) - 2\hat{d}_m^2 \lambda_n^2 }{(\lambda_n + \hat{\zeta}_m)^3 }   \\
		&= \sum_{m=1}^M \frac{ -2(\hat{c}_m-\hat{d}_m)^2 \hat{\zeta}_m^2  }{ (\lambda_n + \hat{\zeta}_m)^3 }  + \sum_{m=1}^M  \frac{2 \hat{d}_m^2 \lambda_n \hat{\zeta}_m }{(\lambda_n + \hat{\zeta}_m)^3 } . \\
	\end{align*}
	From 
	\begin{align*}
		&\sum_{m=1}^M \frac{  (\hat{c}_m-\hat{d}_m)^2 \hat{\zeta}_m^2  }{ (\lambda_n + \hat{\zeta}_m)^3 }    \\
		&\geq \frac{1}{\lambda_n + \hat{\zeta}_M} \sum_{m=1}^M \frac{  (\hat{c}_m-\hat{d}_m)^2 \hat{\zeta}_m^2  }{ (\lambda_n + \hat{\zeta}_m)^2 } \\
		&=\frac{1}{\lambda_n + \hat{\zeta}_M} \|\hat{Z}(\hat{w}^0 - \hat{w}^*) \|_2^2,
	\end{align*}
	we see that, when $\hat{w}^0 \neq \hat{w}^*$, $\sum_{m=1}^M \frac{  (\hat{c}_m-\hat{d}_m)^2 \hat{\zeta}_m^2  }{ (0 + \hat{\zeta}_m)^3 } >0$. So, we have $\hat{\lambda}_n >0$ and $\hat{M}_1(\hat{\lambda}_n) < \hat{M}_1(0)$ when $\hat{w}^0 \neq \hat{w}^*$. 
	
	Let $(\bar{c}_1,...,\bar{c}_M)^{'} = \bar{P}^{'}\bar{w}^0, 
	(\bar{d}_1,...,\bar{d}_M)^{'} = \bar{P}^{'}\hat{w}^* $. Then, we have
	\begin{align*}
		&\bar{M}_1(\lambda_n) \\
		&= \|\bar{Z} \bar{w}^0 - \bar{Z} \hat{w}^* \|_2^2  +\|\bar{Z} \hat{w}^* -  \hat{w}^* \|_2^2  \\
		&=\sum_{m=1}^M \frac{ (\bar{c}_m-\bar{d}_m)^2 \bar{\zeta}_m^2 }{ (\lambda_n + \bar{\zeta}_m)^2 }  + \sum_{m=1}^M \bar{d}_m^2 (\frac{\bar{\zeta}_m}{\lambda_n + \bar{\zeta}_m} -1)^2   \\
		&= \sum_{m=1}^M \frac{ (\bar{c}_m-\bar{d}_m)^2 \bar{\zeta}_m^2 }{ (\lambda_n + \bar{\zeta}_m)^2 }  + \sum_{m=1}^M  \frac{\bar{d}_m^2 \lambda_n^2 }{(\lambda_n + \bar{\zeta}_m)^2 }   ,\\
	\end{align*}
	and
	\begin{align*}
		&\frac{d}{d \lambda_n} \bar{M}_1(\lambda_n) \\
		&= \sum_{m=1}^M \frac{ -2(\bar{c}_m-\bar{d}_m)^2 \bar{\zeta}_m^2  }{ (\lambda_n + \bar{\zeta}_m)^3 }  + \sum_{m=1}^M  \frac{2 \bar{d}_m^2 \lambda_n (\lambda_n + \bar{\zeta}_m) - 2\bar{d}_m^2 \lambda_n^2 }{(\lambda_n + \bar{\zeta}_m)^3 }   \\
		&= \sum_{m=1}^M \frac{ -2(\bar{c}_m-\bar{d}_m)^2 \bar{\zeta}_m^2  }{ (\lambda_n + \bar{\zeta}_m)^3 }  + \sum_{m=1}^M  \frac{2 \bar{d}_m^2 \lambda_n \bar{\zeta}_m }{(\lambda_n + \bar{\zeta}_m)^3 }     . \\
	\end{align*}
	
	Similarly, from
	\begin{align*}
		&\sum_{m=1}^M \frac{  (\bar{c}_m-\bar{d}_m)^2 \bar{\zeta}_m^2  }{ (\lambda_n + \bar{\zeta}_m)^3 }    \\
		&\geq \frac{1}{\lambda_n + \bar{\zeta}_M} \sum_{m=1}^M \frac{  (\bar{c}_m-\bar{d}_m)^2 \bar{\zeta}_m^2  }{ (\lambda_n + \bar{\zeta}_m)^2 } \\
		&=\frac{1}{\lambda_n + \bar{\zeta}_M} \|\bar{Z}(\bar{w}^0 - \hat{w}^*) \|_2^2,
	\end{align*}
	we see that, when $\bar{w}^0 \neq \hat{w}^*$, $\sum_{m=1}^M \frac{  (\bar{c}_m-\bar{d}_m)^2 \bar{\zeta}_m^2  }{ (0 + \bar{\zeta}_m)^3 } >0$. So, we have $\bar{\lambda}_n >0$ and $\bar{M}_1(\bar{\lambda}_n) < \bar{M}_1(0)$ when $\bar{w}^0 \neq \hat{w}^*$. 
	
	\noindent {\bf{Proof of Theorem $\ref{thm:4-3-1}$:}}
	It follows from Lemma $\ref{lem:4}$ and Assumption \ref{ass:4} that
	\begin{align*}
		&E_{S   } [\hat{F}(\hat{w},S ) - \hat{F}(\tilde{w},S)] \\
		&=E_{S   } [\hat{F}(\hat{w} ,S) - \frac{1}{n}C(\hat{w},S) + \frac{1}{n}C(\hat{w},S) - \hat{F}(\tilde{w},S)] \\
		&\leq E_{S   } [\hat{F}(\hat{w} ,S) - \frac{1}{n}C(\hat{w},S) + \frac{1}{n}C(\tilde{w},S) - \hat{F}(\tilde{w},S)] \\
		&= E_{S   }\Big( \frac{2 \sigma^2 \tilde{w}^{'} \kappa }{n} + \frac{\lambda_n \tilde{w}^{'}\tilde{w}  }{n} - \frac{2 \sigma^2 \hat{w}^{'} \kappa }{n}  -\frac{\lambda_n \hat{w}^{'}\hat{w}  }{n}\Big) \\
		&\leq \frac{4 \sigma^2 B_2^{\frac{1}{2}} K M ( 
			1+n^{-2}K^2 )^{\frac{1}{2}}  }{n} + \frac{2 B_2 \lambda_n M  ( 
			1+n^{-2}K^2) }{n} \\
		&=O[n^{-1} \log n K M^2  ( 1+n^{-2}K^2 ) ].
	\end{align*}
	
	Further, from the proof of Lemma $\ref{lem:5}$,
	\begin{align*}
		& E_{S   } [\hat{F}(\bar{w},S) - \hat{F}(\tilde{w},S) ] \\
		&= E_{S   } [\hat{F}(\bar{w},S) - \frac{1}{n}J(\bar{w},S) +  
		\frac{1}{n}J(\bar{w},S) - \hat{F}(\tilde{w},S) ] \\
		&\leq E_{S   } [\hat{F}(\bar{w},S) -  \frac{1}{n}J(\bar{w},S) 
		+  \frac{1}{n}J(\tilde{w},S) - \hat{F}(\tilde{w},S) ] ,
	\end{align*}
	and
	\begin{align*}
		& E_{S   } [\hat{F}(\bar{w},S) -  \frac{1}{n} J(\bar{w},S)  + \frac{\lambda_n \bar{w}^{'}\bar{w}}{n} ] 
		\\
		&=  \frac{1}{n} \sum_{i=1}^n E_{S   } \Big\{  [y_i -x_i^{'} \hat{\theta} (\bar{w}) ]^2  
		-[y_i -x_i^{'} \hat{\theta}^{-i} (\bar{w})] ^2 \Big\} \\
		&=  E_{S   } \Big\{ [y_n -x_n^{'} \hat{\theta} (\bar{w})] ^2  
		-[y_n -x_n^{'} \hat{\theta}^{-n} (\bar{w}) ]^2 \Big\} ,
	\end{align*}
	we have
	$$  E_{S   }\Big \{  [y_n -x_n^{'} \hat{\theta} (\bar{w})] ^2  
	-[y_n -x_n^{'} \hat{\theta}^{-n} (\bar{w})] ^2 \Big\}= O( n^{-1} K^3  
	M^2  ),$$
	that is
	$$  E_{S   } [\hat{F}(\bar{w},S) -  \frac{1}{n} J_S(\bar{w}) ] = O( 
	n^{-1} \log n K^3 M^2  ). $$
	In a similar way, we obtain
	\begin{align*}
		& E_{S   } [\frac{1}{n}J_S(\tilde{w}) - \hat{F}(\tilde{w},S) 
		] = O( n^{-1} \log n  K^3  M^2  ).
	\end{align*}
	So, we have
	$$E_{S   }[\hat{F}(\bar{w},S ) - \hat{F}(\tilde{w},S)] = O(n^{-1} \log 
	n  K^3  M^2 ) .$$
	
	\noindent {\bf{Proof of Theorem $\ref{thm:4-3-2}$:}}
	It follows from Lemmas $\ref{lem:1}$ and $\ref{lem:6}$ that
	\begin{align*}
		&E_{S \sim 
			\mathcal{D}^n} (\|\hat{\theta}(\hat{w}^{-n}) -\hat{\theta}(\hat{w})\|_2^2 ) \\
		&=E_{S  } \Big( \Big\|\sum_{m=1}^M (\hat{w}^{-n}_m-\hat{w}_m) 
		\pi_m \hat{\theta}_m \Big\|_2^2 \Big)\\
		&\leq E_{S  } \Big[\sum_{m=1}^M  \sum_{t=1}^M 
		|(\hat{w}^{-n}_m-\hat{w}_m) (\hat{w}^{-n}_t-\hat{w}_t) 
		\hat{\theta}_m^{'} \pi_m^{'}  \pi_t \hat{\theta}_t| \Big] \\
		&\leq E_{S  } \Big(\sum_{m=1}^M  \sum_{t=1}^M 
		|\hat{w}^{-n}_m-\hat{w}_m| |\hat{w}^{-n}_t-\hat{w}_t| 
		\|\hat{\theta}_m\|_2 \| \hat{\theta}_t\|_2 \Big) \\
		&\leq B_1  K E_{S  } \Big(\sum_{m=1}^M  \sum_{t=1}^M 
		|\hat{w}^{-n}_m-\hat{w}_m| |\hat{w}^{-n}_t-\hat{w}_t| \Big)  \\
		&= B_1  K E_{S  } \Big[ \Big(\sum_{m=1}^M 
		|\hat{w}^{-n}_m-\hat{w}_m| \Big)^2  \Big] \\
		&\leq  B_1 M K  E_{S  }  \Big( \sum_{m=1}^M 
		|\hat{w}^{-n}_m-\hat{w}_m|^2  \Big ) \\
		&=  B_1 M K E_{S  }  [  \|\hat{w}^{-n}-\hat{w} \|_2^2   
		] \\
		&=O [n^{-1} K^4   M^3( 1+n^{-2}K^2)].
	\end{align*}
	Further, it follows from the proof of Lemma $\ref{lem:5}$ that
	\begin{align*}
		&\Big|E_{S  } \Big\{[2y_n -x_n^{'} \hat{\theta}^{-n}(\hat{w}^{-n})- 
		x_n^{'} \hat{\theta}(\hat{w})] [x_n^{'} 
		\hat{\theta}^{-n}(\hat{w}^{-n})-x_n^{'} \hat{\theta}(\hat{w}^{-n}) 
		] \Big \} \Big|  =O[  n^{-1} K^3   M^2( 1+n^{-2}K^2 )] ,
	\end{align*}
	and
	$$|E_{S  } \Big\{ [2y_n -x_n^{'} \hat{\theta}^{-n}(\hat{w}^{-n})- 
	x_n^{'} \hat{\theta}(\hat{w})] [x_n^{'} 
	\hat{\theta}(\hat{w}^{-n})-x_n^{'} \hat{\theta}(\hat{w})] \Big\}| =O[  
	n^{-\frac{1}{2}} K^{\frac{7}{2} }  M^{\frac{5}{2}}( 1+n^{-2}K^2) ] .$$
	From
	\begin{align*}
		&|E_{S  }\Big\{ [ y_n - x_n^{'} \hat{\theta}^{-n}(\hat{w}^{-n})]^2 -  [ y_n - x_n^{'} \hat{\theta}(\hat{w})]^2\Big\}| \\
		&=|E_{S  } \Big\{ [2y_n -x_n^{'} \hat{\theta}^{-n}(\hat{w}^{-n})- x_n^{'} \hat{\theta}(\hat{w})] [x_n^{'} \hat{\theta}^{-n}(\hat{w}^{-n})-x_n^{'} \hat{\theta}(\hat{w})] \Big\}| \\
		&\leq |E_{S  }  \Big\{ [2y_n -x_n^{'} \hat{\theta}^{-n}(\hat{w}^{-n})- x_n^{'} \hat{\theta}(\hat{w})] [x_n^{'} \hat{\theta}^{-n}(\hat{w}^{-n})-x_n^{'} \hat{\theta}(\hat{w}^{-n})] \Big\}|\\
		&\ \ \ +|E_{S  } \Big\{ [2y_n -x_n^{'} \hat{\theta}^{-n}(\hat{w}^{-n})- x_n^{'} \hat{\theta}(\hat{w})] [x_n^{'} \hat{\theta}(\hat{w}^{-n})-x_n^{'} 
		\hat{\theta}(\hat{w})]\Big\}|,
	\end{align*}
	we have
	$$E_{S  }\Big\{ [ y_n - x_n^{'} \hat{\theta}^{-n}(\hat{w}^{-n})]^2 -  [ 
	y_n - x_n^{'} \hat{\theta}(\hat{w})]^2\Big\} =O [n^{-\frac{1}{2}} 
	K^{\frac{7}{2} }  M^{\frac{5}{2}}( 1+n^{-2}K^2 )].   $$
	In a similar way, we obtain
	$$E_{S  ,z^*  } 
	\Big\{[ y^* - x^{*'} \hat{\theta}^{-n}(\hat{w}^{-n})] ^2  -  
	[y^* - x^{*'} \hat{\theta}(\hat{w})] ^2 \Big\} =O [n^{-\frac{1}{2}} 
	K^{\frac{7}{2} }  M^{\frac{5}{2}}( 1+n^{-2}K^2 ) ].$$
	
	On the other hand, it follows from Lemmas $\ref{lem:1}$ and $\ref{lem:6}$ that
	\begin{align*}
		&E_{S \sim 
			\mathcal{D}^n}[\|\hat{\theta}(\bar{w}^{-n}) -\hat{\theta}(\bar{w})\|_2^2] \\
		&=E_{S  } \Big[\Big\|\sum_{m=1}^M (\bar{w}^{-n}_m-\bar{w}_m) 
		\pi_m \hat{\theta}_m\Big\|_2^2 \Big]\\
		&\leq E_{S  } \Big[\sum_{m=1}^M  \sum_{t=1}^M 
		|(\bar{w}^{-n}_m-\bar{w}_m) (\bar{w}^{-n}_t-\bar{w}_t) 
		\hat{\theta}_m^{'} \pi_m^{'}  \pi_t \hat{\theta}_t| \Big] \\
		&\leq E_{S  } \Big(\sum_{m=1}^M  \sum_{t=1}^M 
		|\bar{w}^{-n}_m-\bar{w}_m| |\bar{w}^{-n}_t-\bar{w}_t| 
		\|\hat{\theta}_m\|_2 \| \hat{\theta}_t\|_2\Big)  \\
		&\leq B_1  K E_{S  } \Big(\sum_{m=1}^M  \sum_{t=1}^M 
		|\bar{w}^{-n}_m-\bar{w}_m| |\bar{w}^{-n}_t-\bar{w}_t|  \Big) \\
		&= B_1  K E_{S  } \Big[ \Big(\sum_{m=1}^M 
		|\bar{w}^{-n}_m-\bar{w}_m| \Big)^2  \Big] \\
		&\leq  B_1 M K  E_{S  }  \Big( \sum_{m=1}^M 
		|\bar{w}^{-n}_m-\bar{w}_m|^2  \Big)  \\
		&=  B_1 M K E_{S  }   ( \|\bar{w}^{-n}-\bar{w} \|_2^2 )  
		\\
		&=O( n^{-1} K^4   M^3) .
	\end{align*}
	So, from the proof of Lemma $\ref{lem:5}$, we have
	\begin{align*}
		&|E_{S  } \Big\{ [2y_n -x_n^{'} \hat{\theta}^{-n}(\bar{w}^{-n})- 
		x_n^{'} \hat{\theta}(\bar{w})] [x_n^{'} 
		\hat{\theta}^{-n}(\bar{w}^{-n})-x_n^{'} \hat{\theta}(\bar{w}^{-n})] 
		\Big\}|  =O(  n^{-1} K^3   M^2) ,
	\end{align*}
	and
	$$|E_{S  } \Big\{ [2y_n -x_n^{'} \hat{\theta}^{-n}(\bar{w}^{-n})- 
	x_n^{'} \hat{\theta}(\bar{w})] [x_n^{'} 
	\hat{\theta}(\bar{w}^{-n})-x_n^{'} \hat{\theta}(\bar{w})] \Big\}| =O(  
	n^{-\frac{1}{2}} K^{\frac{7}{2} }  M^{\frac{5}{2}}) .$$
	From
	\begin{align*}
		&|E_{S  }\Big\{ [ y_n - x_n^{'} \hat{\theta}^{-n}(\bar{w}^{-n})]^2 -  [ y_n - x_n^{'} \hat{\theta}(\bar{w})]^2\Big\}| \\
		&=|E_{S  } \Big\{ [2y_n -x_n^{'} \hat{\theta}^{-n}(\bar{w}^{-n})- x_n^{'} \hat{\theta}(\bar{w})][ x_n^{'} \hat{\theta}^{-n}(\bar{w}^{-n})-x_n^{'} \hat{\theta}(\bar{w})] \Big\}| \\
		&\leq |E_{S  } \Big\{ [2y_n -x_n^{'} \hat{\theta}^{-n}(\bar{w}^{-n})- x_n^{'} \hat{\theta}(\bar{w})] [x_n^{'} \hat{\theta}^{-n}(\bar{w}^{-n})-x_n^{'} \hat{\theta}(\bar{w}^{-n})] \Big\}|\\
		&+|E_{S  }\Big\{ [2y_n -x_n^{'} \hat{\theta}^{-n}(\bar{w}^{-n})- x_n^{'} \hat{\theta}(\bar{w})] [x_n^{'} \hat{\theta}(\bar{w}^{-n})-x_n^{'} 
		\hat{\theta}(\bar{w})] \Big\}|,
	\end{align*}
	we know that
	$$E_{S  }\Big\{ [ y_n - x_n^{'} \hat{\theta}^{-n}(\bar{w}^{-n})]^2 -  [ 
	y_n - x_n^{'} \hat{\theta}(\bar{w})]^2\Big\} =O( n^{-\frac{1}{2}} 
	K^{\frac{7}{2} }  M^{\frac{5}{2}}).   $$
	In a similar way, we obtain
	$$E_{S  ,z^*  } \Big\{ 
	[y^* - x^{*'} \hat{\theta}^{-n}(\bar{w}^{-n}) ]^2  -  
	[y^* - x^{*'} \hat{\theta}(\bar{w})] ^2 \Big\} =O( n^{-\frac{1}{2}} 
	K^{\frac{7}{2} }  M^{\frac{5}{2}}) .$$
	
	Similarly, it follows from Lemmas $\ref{lem:1}$ and $\ref{lem:6}$ that
	\begin{align*}
		&E_{S \sim 
			\mathcal{D}^n}[\|\hat{\theta}(\hat{w}^{*,-n}) -\hat{\theta}(\hat{w}^*)\|_2^2] \\
		&=E_{S  } \Big[\Big\|\sum_{m=1}^M (\hat{w}^{*,-n}_m-\hat{w}^*_m) 
		\pi_m \hat{\theta}_m\Big\|_2^2\Big] \\
		&\leq E_{S  } \Big[\sum_{m=1}^M  \sum_{t=1}^M 
		|(\hat{w}^{*,-n}_m-\hat{w}^*_m) (\hat{w}^{*,-n}_t-\hat{w}^*_t) 
		\hat{\theta}_m^{'} \pi_m^{'}  \pi_t \hat{\theta}_t| \Big] \\
		&\leq E_{S  } [\sum_{m=1}^M  \sum_{t=1}^M 
		|\hat{w}^{*,-n}_m-\hat{w}^*_m| |\hat{w}^{*,-n}_t-\hat{w}^*_t| 
		\|\hat{\theta}_m\|_2 \| \hat{\theta}_t\|_2 ] \\
		&\leq B_1  K E_{S  } \Big(\sum_{m=1}^M  \sum_{t=1}^M 
		|\hat{w}^{*,-n}_m-\hat{w}^*_m| |\hat{w}^{*,-n}_t-\hat{w}^*_t|\Big)   \\
		&= B_1  K E_{S  } \Big[ \Big(\sum_{m=1}^M 
		|\hat{w}^{*,-n}_m-\hat{w}^*_m| \Big)^2  \Big] \\
		&\leq  B_1 M K  E_{S  }   \Big(\sum_{m=1}^M 
		|\hat{w}^{*,-n}_m-\hat{w}^*_m|^2   \Big) \\
		&=  B_1 M K E_{S  }    (\|\hat{w}^{*,-n}-\hat{w}^* \|_2^2 )  
		\\
		&=O( n^{-1} K^4   M^4 ) .
	\end{align*}
	Thus, from the proof of Lemma $\ref{lem:5}$, we have
	\begin{align*}
		&|E_{S  } \Big\{ [2y_n -x_n^{'} \hat{\theta}^{-n}(\hat{w}^{*,-n})- 
		x_n^{'} \hat{\theta}(\hat{w}^*)] [x_n^{'} 
		\hat{\theta}^{-n}(\hat{w}^{*,-n})-x_n^{'} 
		\hat{\theta}(\hat{w}^{*,-n})] \Big\}|  =O(  n^{-1} K^3   M^{3}) ,
	\end{align*}
	and
	$$|E_{S  }\Big\{ [2y_n -x_n^{'} \hat{\theta}^{-n}(\hat{w}^{*,-n})- 
	x_n^{'} \hat{\theta}(\hat{w}^*)] [x_n^{'} 
	\hat{\theta}(\hat{w}^{*,-n})-x_n^{'} \hat{\theta}(\hat{w}^*)] \Big\}| 
	=O(  n^{-\frac{1}{2}} K^{\frac{7}{2} }  M^{\frac{7}{2}}).$$
	From
	\begin{align*}
		&|E_{S  }\Big\{ [ y_n - x_n^{'} \hat{\theta}^{-n}(\hat{w}^{*,-n})]^2 -  [ y_n - x_n^{'} \hat{\theta}(\hat{w}^*)]^2\Big\}| \\
		&=|E_{S  }\Big\{ [2y_n -x_n^{'} \hat{\theta}^{-n}(\hat{w}^{*,-n})- x_n^{'} \hat{\theta}(\hat{w}^*)] [x_n^{'} \hat{\theta}^{-n}(\hat{w}^{*,-n})-x_n^{'} \hat{\theta}(\hat{w}^*)]  \Big\}| \\
		&\leq |E_{S  }\Big\{ (2y_n -x_n^{'} \hat{\theta}^{-n}(\hat{w}^{*,-n})- x_n^{'} \hat{\theta}(\hat{w}^*)] [x_n^{'} \hat{\theta}^{-n}(\hat{w}^{*,-n})-x_n^{'} \hat{\theta}(\hat{w}^{*,-n})]  \Big\}|\\
		&+|E_{S  }\Big\{ [2y_n -x_n^{'} \hat{\theta}^{-n}(\hat{w}^{*,-n})- x_n^{'} \hat{\theta}(\hat{w}^*)] [x_n^{'} \hat{\theta}(\hat{w}^{*,-n})-x_n^{'} 
		\hat{\theta}(\hat{w}^*)]  \Big\}|,
	\end{align*}
	we know that
	$$E_{S  }\Big\{ [ y_n - x_n^{'} \hat{\theta}^{-n}(\hat{w}^{*,-n})]^2 -  
	[ y_n - x_n^{'} \hat{\theta}(\hat{w}^*)]^2\Big\} =O( n^{-\frac{1}{2}} 
	K^{\frac{7}{2} }  M^{\frac{7}{2}}).   $$
	In a similar way, we obtain
	$$E_{S  ,z^*  } 
	\Big\{ [y^* - x^{*'} \hat{\theta}^{-n}(\hat{w}^{*,-n}) ]^2  -  
	[y^* - x^{*'} \hat{\theta}(\hat{w}^*)] ^2 \Big\} =O( n^{-\frac{1}{2}} 
	K^{\frac{7}{2} }  M^{\frac{7}{2}}) .$$



\bibliographystyle{elsarticle-harv} 
\bibliography{ref}
\addcontentsline{toc}{section}{References}





\end{document}